\documentclass[opre, nonblindrev]{informs3}
\DoubleSpacedXII
\usepackage{natbib}
\bibpunct[, ]{(}{)}{,}{a}{}{,}%
\def\bibfont{\small}%

\TheoremsNumberedThrough
\EquationsNumberedThrough

\usepackage{amsmath, comment, tikz, graphicx, algorithm, algpseudocode, multirow, multicol, subcaption, bbm, epsfig, enumerate, amsfonts, enumitem, subeqnarray, url, color, array, xcolor, booktabs, endnotes, float, mathpazo, pdfpages, bm, pdflscape}
\usepackage[title]{appendix}
\usepackage{diagbox}
\usepackage{threeparttable}
\usepackage{titlesec}
\usepackage[resetlabels]{multibib}
\newcites{ec}{References} 

\titleformat{\section}{\centering\normalfont\rmfamily\fontsize{13pt}{26pt}\selectfont\bfseries}
  {\thesection}{1em}{}

\titleformat{\subsection}{\centering\normalfont\rmfamily\normalsize\bfseries}  {\thesubsection}{1em}{}

\titlespacing{\subsubsection}
    {0pt}{9pt}{6pt}

\titleformat{\subsubsection}{\normalfont\rmfamily\normalsize\bfseries}  {\thesubsubsection}{1em}{}

\makeatletter
\newcommand{\otherlabel}[2]{\protected@edef\@currentlabel{#2}\label{#1}}
\makeatother

\usetikzlibrary{decorations.pathreplacing}
\usetikzlibrary{math}
\usepackage[skip=2pt, font=footnotesize, labelfont=bf, labelsep=period]{caption}

\usepackage{pgfplots}
    \pgfplotsset{
        compat=1.9,
    }
\usepackage[backref = false, bookmarks, breaklinks=true, colorlinks = true, plainpages = false, citecolor = DarkBlue , urlcolor = DarkBlue, filecolor = DarkBlue, linkcolor = DarkBlue]{hyperref}
\definecolor{DarkBlue}{rgb}{0,0.08,0.45}
\usepackage{tablefootnote}

\let\footnote=\endnote
\let\enotesize=\footnotesize
\def\notesname{Endnotes}%
\def\makeenmark{$^{\theenmark}$}
\def\enoteformat{\rightskip0pt\leftskip0pt\parindent=1.75em
  \leavevmode\llap{\theenmark.\enskip}\baselineskip10pt}

\renewcommand{\theRRHFirstLine}{}
\renewcommand{\theRRHSecondLine}{}
\renewcommand{\theLRHFirstLine}{}
\renewcommand{\theLRHSecondLine}{}

\usepackage{CJKutf8}

\usepackage[backref = false, bookmarks]{hyperref}
\hypersetup{hidelinks}

\renewcommand{\theRRHFirstLine}{}
\renewcommand{\theRRHSecondLine}{}
\renewcommand{\theLRHFirstLine}{}
\renewcommand{\theLRHSecondLine}{}

\algdef{SE}[DOWHILE]{Do}{doWhile}{\algorithmicdo}[1]{\algorithmicwhile\ #1}%

\usepackage{titlesec}

\ECRepeatTheorems
\EquationsNumberedThrough

\begin{document}


\TITLE{A General Method for Detecting Information Generated by Large Language Models }

\ARTICLEAUTHORS{
\AUTHOR{Minjia Mao*}
\AFF{University of Delaware;
\EMAIL{mjmao@udel.edu}}
\AUTHOR{Donjun Wei*}
\AFF{The University of Hong Kong; 
\EMAIL{dongjun@connect.hku.hk}}
\AUTHOR{Xiao Fang}
\AFF{University of Delaware;
\EMAIL{xfang@udel.edu}}
\AUTHOR{Michael Chau}
\AFF{The University of Hong Kong; 
\EMAIL{mchau@business.hku.hk}}

$^{*}$Equal Contributions
}

\OneAndAHalfSpacedXII
\ABSTRACT{\normalsize\textbf{Abstract:} The proliferation of large language models (LLMs) has significantly transformed the digital information landscape, making it increasingly challenging to distinguish between human-written and LLM-generated content. Detecting LLM-generated information is essential for preserving trust on digital platforms (e.g., social media and e-commerce sites) and preventing the spread of misinformation, a topic that has garnered significant attention in IS research. However, current detection methods, which primarily focus on identifying content generated by specific LLMs in known domains, face challenges in generalizing to new (i.e., unseen) LLMs and domains. This limitation reduces their effectiveness in real-world applications, where the number of LLMs is rapidly multiplying and content spans a vast array of domains. In response, we introduce a general LLM detector (GLD) that combines a twin memory networks design and a theory-guided detection generalization module to detect LLM-generated information across unseen LLMs and domains. Using real-world datasets, we conduct extensive empirical evaluations and case studies to demonstrate the superiority of GLD over state-of-the-art detection methods. The study has important academic and practical implications for digital platforms and LLMs.}%


\maketitle

\DoubleSpacedXII






\newpage
\section{INTRODUCTION}

Large language models (LLMs) are artificial intelligence (AI) systems designed to understand human languages and generate human-like content \citep{Ouyang0JAWMZASR22}. In recent years, the number of LLMs has grown exponentially, driven by both open-source communities and commercial companies. For example, Hugging Face, a platform hosting open-source LLMs, provides access to over 100,000 publicly available LLMs, with new ones added each month.\SingleSpacedXII\footnotemark\footnotetext{See \url{https://huggingface.co/blog/2023-in-llms} (last accessed on June 15, 2025)}\DoubleSpacedXII~Leading technology companies, such as OpenAI and Cohere, roll out new versions of commercial LLMs every few months. The capabilities of LLMs like OpenAI's GPT-4o and Meta's LLaMA-3 have reached remarkable levels, enabling them to generate human-like content in seconds \citep{mousavi2020harnessing,reisenbichler2022frontiers,goli2024frontiers}. While the rapid advancement of LLMs has resulted in many useful applications, it has also fueled the creation of fake and misleading information online \citep{berente2021managing,fang2024bias}.
For instance, social media platforms like X (formerly known as Twitter) and Facebook now contain an increasing amount of LLM-generated information across various domains, such as news and product reviews.\SingleSpacedXII\footnotemark\footnotetext{ See \url{https://nymag.com/intelligencer/article/ai-generated-content-internet-online-slop-spam.html} (last accessed on June 15, 2025)}\DoubleSpacedXII~
Similarly, a recent investigation reveals that over 40\% of articles on Medium, a prominent U.S. blogging platform, are generated by LLMs without fact-checking or proper editing, covering diverse domains such as personal stories, reports, and news.\SingleSpacedXII\footnotemark\footnotetext{See \url{https://www.wired.com/story/ai-generated-medium-posts-content-moderation} (last accessed on June 15, 2025)}\DoubleSpacedXII

The enormous and rapidly growing amount of information produced by LLMs could impose serious harm on both information consumers and providers (e.g., digital platforms) \citep{pennycook2021psychology}. 
First, since LLMs can produce highly human-like content, it is becoming increasingly difficult for information consumers to distinguish between LLM-generated and human-written information, potentially exposing themselves to economic losses \citep{wei2022combining,lee2024disinformation}.
For example, click farms now leverage LLMs to create human-like posts across various domains, often including deceptive links to mislead social media users into purchasing counterfeit or low-quality products.\SingleSpacedXII\footnotemark\footnotetext{See \url{https://www.404media.co/facebooks-algorithm-is-boosting-ai-spam-that-links-to-ai-generated-ad-laden-click-farms} (last accessed on June 15, 2025)}\DoubleSpacedXII~
Second, the growing volume of LLM-generated information undermines consumers' trust in information providers. 
%
%
For example, overwhelmed by blogs generated by various LLMs, Medium struggles to provide quality information to its users, resulting in users leaving the platform.\SingleSpacedXII\footnotemark\footnotetext{See \url{https://reutersinstitute.politics.ox.ac.uk/news/ai-generated-slop-quietly-conquering-internet-it-threat-journalism-or-problem-will-fix-itself} (last accessed on June 15, 2025)}\DoubleSpacedXII~
Moreover, the harm caused by LLM-generated information extends beyond the information providers who publish it. LLM-generated information could impact search engine rankings, thereby overshadowing high-quality human-written content sought by users.\SingleSpacedXII\footnotemark\footnotetext{See \url{https://www.theverge.com/2024/7/31/24210304/wix-ai-website-builder-seo-blog-posts} (last accessed on June 15, 2025)}\DoubleSpacedXII~As a result, the entire information ecosystem suffers from the proliferation of LLM-generated information. 
To protect information consumers and providers and preserve the integrity of the information ecosystem, it is critical to design a method that can accurately detect information generated by various LLMs across diverse domains. 
The status quo for detecting LLM-generated information can be classified into two categories: zero-shot and learning-based methods. 
Zero-shot methods identify LLM-generated information without the need for training on labeled datasets. Instead, they employ a pre-trained LLM, known as a scoring LLM, to score each piece of information and then distinguish human-written information from LLM-generated information based on their scores. For example, DetectGPT, a prominent zero-shot method, perturbs randomly selected tokens in a piece of information and then utilizes a scoring LLM to calculate the perplexity\SingleSpacedXII\footnotemark\footnotetext{Given a piece of information, perplexity is the exponential of the average negative log likelihood of each word in the information.}\DoubleSpacedXII of the information before and after the perturbation \citep{mitchell2023detectgpt}. An increase in perplexity after the perturbation suggests that the information is likely generated by an LLM.
Zero-shot methods are effective in detecting information generated by the same LLM as the scoring LLM but are ineffective when the scoring LLM differs from the one used to generate the information \citep{liu2023survey}. 
Different LLMs vary in model architecture, size, and training approach. As a result, a scoring LLM cannot reliably evaluate the information generated by a different LLM, leading to poor detection performance \citep{su2023detectllm}.
If an LLM that generates information to be detected differs from the scoring LLM utilized by a zero-shot method, we refer to the information-generating LLM as an \textbf{\textit{unseen LLM}} for the method. Zero-shot methods are ineffective in detecting information produced by unseen LLMs \citep{liu2023survey}.

Learning-based methods are trained on data comprising both human-written content and information generated by specific LLMs (e.g., ChatGPT and LLaMA) in particular domains (e.g., product reviews and news)~\citep{gehrmann2019gltr,guo2023close}. For example, training data might contain human-written and ChatGPT-generated product reviews as well as human-written and LLaMA-generated news. 
These methods are trained to detect information produced by the same LLMs in the same domains as those in their training data, 
thereby generalizing poorly to other LLMs and domains \citep{frohling2021feature,mitchell2023detectgpt}.
As a result, learning-based methods are ineffective in detecting information generated by LLMs in domains different from those in their training data \citep{Li2023MAGEMT}, which we refer to as \textbf{\textit{unseen LLMs}} and \textbf{\textit{unseen domains}}. 
Given the vast number of LLMs and the wide range of domains, existing learning-based methods can ``see'' only a few LLMs and domains --- that is, they are trained on information generated by a small number of LLMs across several domains. 
Consequently, these methods perform poorly in detecting information generated by a large number of unseen LLMs in numerous unseen domains. 
Therefore, it is necessary to develop a general method that is trained on information produced by a few LLMs in several domains but can generalize to \textit{\textbf{identify information produced by unseen LLMs in unseen domains effectively}}.
In response to the aforementioned research gap, we propose a general LLM detector (GLD) to distinguish between human-written and LLM-generated information. Compared to existing methods for detecting LLM-generated content, the key advantage of GLD is its ability to generalize to unseen LLMs and domains. To achieve this advantage, our study introduces two methodological innovations.
First, for each piece of information, it is essential to extract and represent features of the domain to which it belongs, as well as features of its author (e.g., human or an LLM). 
To this end, we propose a Twin Memory Networks  (TMN) design, which learns the embeddings of the domain, author, and textual content for each piece of information. 
Second, we theoretically analyze the generalizability of a detection method to unseen LLMs and domains. Built on the theoretical analysis, we propose a Detection Generalization Module (DGM) that can effectively generalize to unseen LLMs and domains by learning LLM- and domain-invariant embeddings.
Empirically, we evaluate our GLD method on a dataset that combines human-written information and information generated by multiple LLMs, such as GPT-4o and LLaMA-3, in multiple domains, including news and online reviews. The empirical results demonstrate that our proposed GLD method achieves state-of-the-art detection performance compared to benchmark methods, including both zero-shot and learning-based methods, for unseen LLMs and domains. In addition, we conduct case studies in two unseen domains, namely online reviews and academic writing, to demonstrate the practicality of the proposed method in real-world applications that are highly relevant to IS research.

\section{LITERATURE REVIEW}
\label{sec:rw}

Existing methods for detecting LLM-generated information can be classified into two categories: zero-shot and learning-based methods. Learning-based methods can be further divided into feature-based and fine-tuning methods. In this section, we review representative methods in each category and highlight the key novelties of our study.


\subsection{Zero-shot Methods}

Zero-shot methods identify LLM-generated information without requiring training on labeled datasets. These methods utilize a pre-trained LLM to score each piece of information and differentiate human-written content from LLM-generated information on the basis of their scores. 
DetectGPT, a representative zero-shot method, perturbs (i.e., rewrites) randomly selected tokens in a piece of information and then employs a GPT model to evaluate its perplexity before and after the perturbation \citep{mitchell2023detectgpt}. A consistent increase in perplexity across multiple perturbations indicates that the information is likely LLM-generated.
Fast-DetectGPT improves the efficiency of DetectGPT through sampling, while maintaining detection performance \citep{bao2023fast}. \cite{su2023detectllm} employs a scoring LLM to compute the likelihood of each token in a piece of information, given its previous tokens. They then define a ratio between the token's log-likelihood and its log-rank. They observe that the average ratio across tokens in human-written content is greater than that in LLM-generated information, making the ratio a useful metric to distinguish between them.

By utilizing a pre-trained LLM as the scoring model, a zero-shot method can effectively detect content generated by the same LLM as the scoring LLM. 
For example, when equipped with GPT-J as the scoring LLM, DetectGPT achieves an AUC of 0.92 in detecting GPT-J-generated information across various domains \citep{mitchell2023detectgpt}. 
However, relying on a particular scoring LLM limits the generalizability of a zero-shot method in detecting information generated by unseen LLMs that are different from the scoring LLM.
For instance, when using with GPT-J as the scoring LLM, DetectGPT struggles to identify information generated by GPT-2, attaining an AUC of only 0.6 \citep{mitchell2023detectgpt}. 

\subsection{Feature-based and Fine-tuning Methods}

Feature-based and fine-tuning methods are learning-based approaches trained using labeled data that include human-written content and information generated by specific LLMs in particular domains. Specifically, feature-based methods extract probability-based or linguistic features from training datasets. The extracted features and their corresponding labels are then used to train a classifier to distinguish between human-written and LLM-generated information. For example, GLTR utilizes GPT-2 to compute the likelihood rank of each token in a piece of information \citep{gehrmann2019gltr}. Compared to human-written information, LLM-generated information typically contains a higher frequency of high-ranked tokens. Therefore, a logistic regression model can be trained on the frequency of high-ranked tokens to differentiate between GPT-2-generated and human-written information. Perplexity, another probability-based feature, has also been used to identify LLM-generated information. Since LLM-generated information usually has lower perplexity than human-written content, a classifier can be built on it to detect LLM-generated information \citep{guo2023close,wu2023survey}.
Meanwhile, as an exemplar method leveraging linguistic features, \cite{frohling2021feature} extract such features as readability and lexical diversity from a labeled dataset and then train a logistic regression model to identify GPT-generated information. 

While feature-based methods perform well at detecting information generated by the same LLMs in the same domains as those in
their training datasets, their performance deteriorates significantly when detecting information produced by unseen LLMs and domains different from those in the training data \citep{gehrmann2019gltr}. 
Probability-based features, such as token likelihood and perplexity, depend on the specific LLM used to calculate them \citep{wu2023survey}. Consequently, their values vary across different LLMs.
Similarly, linguistic features, such as readability and repetitiveness, differ across content generated by various LLMs. For instance, the readability score of information generated by ChatGPT is significantly different from that of information produced by GPT-2 \citep{Li2023MAGEMT}. 
Moreover, linguistic features like named entities and lexical diversity vary significantly across different domains \citep{liu2021crossner, guo2024benchmarking}. Because of the reliance of probability-based and linguistic features on specific LLMs and domains, detection methods trained on these features generalize poorly to unseen LLMs and domains.


Unlike feature-based methods, fine-tuning methods do not extract features from training datasets. Instead, they fine-tune a transformer-based model, such as BERT \citep{devlin2019bert}, RoBERTa \citep{liu2019roberta}, and Longformer \citep{beltagy2020longformer}, on these datasets to classify information as either LLM-generated or human-written.
For example, OpenAI introduces the GPT-2 output detector \citep{solaiman2019release}, which is implemented by fine-tuning RoBERTa on a dataset comprising both GPT-2-generated and human-written content. 
More recently, \citet{liu2023argugpt} fine-tune RoBERTa to detect GPT-generated English essays and demonstrate that it outperforms zero-shot and feature-based methods.
However, fine-tuning methods are prone to overfitting their training datasets, thereby generalizing poorly to unseen LLMs and domains that differ from those in the training data \citep{mitchell2023detectgpt}. 
For example, OpenAI's GPT-2 output detector fails to identify information generated by its unseen LLMs, such as ChatGPT \citep{guo2023close}. 
As another example, \citet{Li2023MAGEMT} show that a method that fine-tunes the Longformer model on a training dataset of human-written content across 10 domains, along with corresponding texts generated by 27 LLMs, still struggles to detect information generated by an unseen LLM in an unseen domain.
\subsection{Key Novelties of Our Method}
As discussed above, existing zero-shot methods rely on a specific scoring LLM, making them ineffective in detecting information generated by unseen LLMs.
Existing learning-based methods are trained on datasets comprising human-written texts and information generated by specific LLMs in particular domains. As a result, while these methods are good at detecting information generated by the same LLMs in the same domains as those in their training datasets, they generalize poorly to unseen LLMs and new domains.

Unlike existing methods, our proposed method can generalize to detect information generated by unseen LLMs in new domains effectively. This capability of our method is realized through two key methodological innovations. 
First, to generalize to unseen LLMs and domains, it is crucial to extract and represent domain and author features for each piece of information. Therefore, different from existing methods that focus solely on learning the embedding of the textual content for each piece of information \citep{solaiman2019release,bhattacharjee2024eagle}, our method also learns the embeddings of its domain and author, in addition to the textual content.
Second, we conduct a rigorous theoretical analysis of the generalizability of a detection method to unseen LLMs and domains. Based on the analysis, we propose to detect LLM-generated information by learning LLM- and domain-invariant embeddings for LLM-generated information and domain-invariant embeddings for human-written information.
\section{PROBLEM DEFINITION}
\label{sec:pd}

We consider the problem of detecting LLM-generated information in a general manner. As the number of LLMs grows rapidly and they are deployed to generate textual information across an increasingly wide range of domains, it is essential to develop a detection method that can generalize effectively to unseen LLMs and new domains, 
by training on documents \SingleSpacedXII\footnotemark\footnotetext{We use the term ``document'' to refer to an instance of textual information collected in the dataset, such as a news article, a review, or a story.}\DoubleSpacedXII~written by humans or generated by a few LLMs in a few domains.
Concretely, let $\{H\} \cup G$ represent the authors of the collected documents, where $H$ denotes humans and $G = \{G_1, G_2, \dots, G_m\}$ denotes the set of $m$ LLMs. Let $S=\{S_{1},S_{2},\cdots,S_{n}\}$ represent the set of $n$ domains to which the collected documents belong. The training corpus consists of the collected $N$ documents, each annotated with its author $g_k \in {H} \cup G$, domain $s_k \in S$, and label $y_k \in \{0, 1\}$, where $y_k = 0$ indicates a human-written document and $y_k = 1$ indicates one generated by an LLM, $k = 1, 2, \dots, N$.
We now formally define the general LLM-generated information detection problem as follows.

\textbf{Definition 1 (General LLM-generated Information Detection Problem (GLID)).}
A training corpus contains $N$ documents, each of which is written by humans $H$ or generated by one of the $m$ LLMs, denoted by $G=\{G_{1},G_{2},\cdots,G_{m}\}$, and belongs to one of the $n$ domains, denoted by $S=\{S_{1},S_{2},\cdots,S_{n}\}$. Accordingly, each training document $d_k$ is labeled as human-written $(y_k=0)$ or LLM-generated $(y_k=1)$, where $k=1,2,\dots,N$. It is also annotated with its domain $s_k \in S$ and author $g_k \in {H} \cup G$. The objective is to learn a model from the training corpus to classify a test document written by humans or generated by any unseen LLMs (not in $G$) in any unseen domains (not in $S$), as human-written or LLM-generated. 
\section{METHOD}
\label{sec:method}

\begin{figure}[htbp]
\FIGURE{\includegraphics[width=0.95\textwidth]{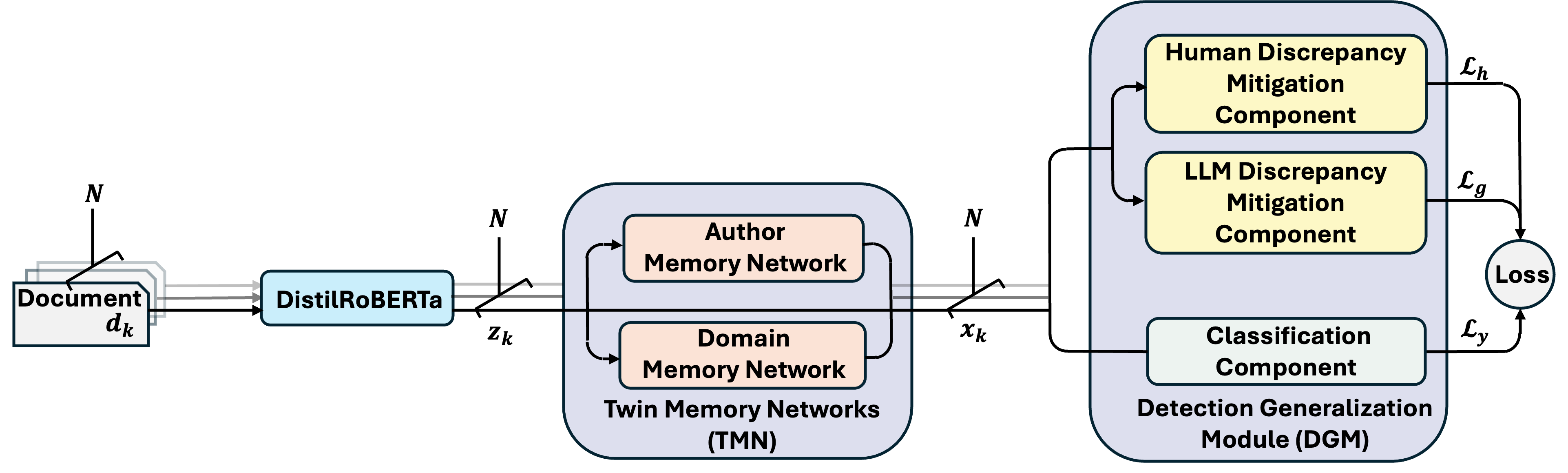}}
{Overall Architecture of General LLM Detector (GLD). 
\label{fig:method-overview}}
{}
\end{figure}

We propose a general LLM detector (GLD) to solve the GLID problem. The overall architecture of the GLD method is shown in Figure~\ref{fig:method-overview}. As illustrated, for each training document, GLD leverages a pre-trained model to represent its textual content as a textual embedding. 
The textual embedding is then passed to the Twin Memory Networks module of GLD, which comprises two memory networks with similar architectures. These networks extract the author and domain embeddings from the document, which are then combined with the textual embedding to form the final representation of the document.
Using the final embeddings of training documents as input, the Detection Generalization Module learns a detector capable of effectively identifying information generated by unseen LLMs in new domains.
We present the design of each module in Sections~\ref{sec:method-tmn}  and ~\ref{sec:method-dmg}, respectively. Section~\ref{sec:method-training} describes the training process of GLD and how it is used to classify a test document as LLM-generated or human-written. For the convenience of the
reader, we summarize important notations in Table~\ref{tab:notations}. 

\begin{table}[htbp]
\centering
\OneAndAHalfSpacedXI
\caption{Notation.}\label{tab:notations}
\def\arraystretch{1.3}
\begin{tabular}{ p{2cm} p{13cm}  }
\toprule
Notation & Description \\
\midrule
$H$ & Humans \\	
$G_i$ & An LLM, $i=1,2,\dots,m$ \\
$S_j$ & A domain, $j=1,2,\dots,n$ \\
$d_k$ & A document, $k=1,2,\dots,N$ \\ 
$g_k$ & The author of document $d_k$, $g_k \in \{H, G_1, G_2,\dots,G_m\}$ \\
$s_k$ & The domain to which document $d_k$ belongs, $s_k \in \{S_1, S_2,\dots,S_n\}$ \\ 
$y_k$ & The label of document $d_k$, $y_k=0$ (human-written) or 1 (LLM-generated) \\
$z_k$ & The textual embedding of document $d_k$ \\
$z_k^g$, $z_k^s$ & The document-specific author and domain embeddings of $d_k$ \\
$x_k$ & The final embedding of document $d_k$, which is the concatenation of $z_k$, $z_k^g$, and $z_k^s$   \\
$\mathcal{D}_{H,S_j}$ & The set of final embeddings corresponding to documents written by humans $H$ in domain $S_j$ \\ 
$\mathcal{D}_{G_i,S_j}$ & The set of final embeddings corresponding to documents generated by LLM $G_i$ in domain $S_j$ \\ 
$\mathcal{D}_{ij}$ & A labeled dataset, $\mathcal{D}_{ij}=\mathcal{D}_{H,S_j}\cup \mathcal{D}_{G_i,S_j}$ \\ 
$\mathbb{D}_{ij}$ & The distribution of document embeddings in $\mathcal{D}_{ij}$ \\ 
$\mathbb{D}_u$ & The distribution of embeddings corresponding to documents generated by unseen LLMs or written by humans in unseen domains \\ 
$\mathcal{L}_h$ & Discrepancy mitigation loss (Human)\\ 
$\mathcal{L}_g$ & Discrepancy mitigation loss (LLM) \\ 
$\mathcal{L}_y$ & Classification loss \\ 
\bottomrule
\end{tabular}
\end{table}

\subsection{Twin Memory Networks} \label{sec:method-tmn}

\begin{figure}[htbp]
\FIGURE{\includegraphics[width=\textwidth]{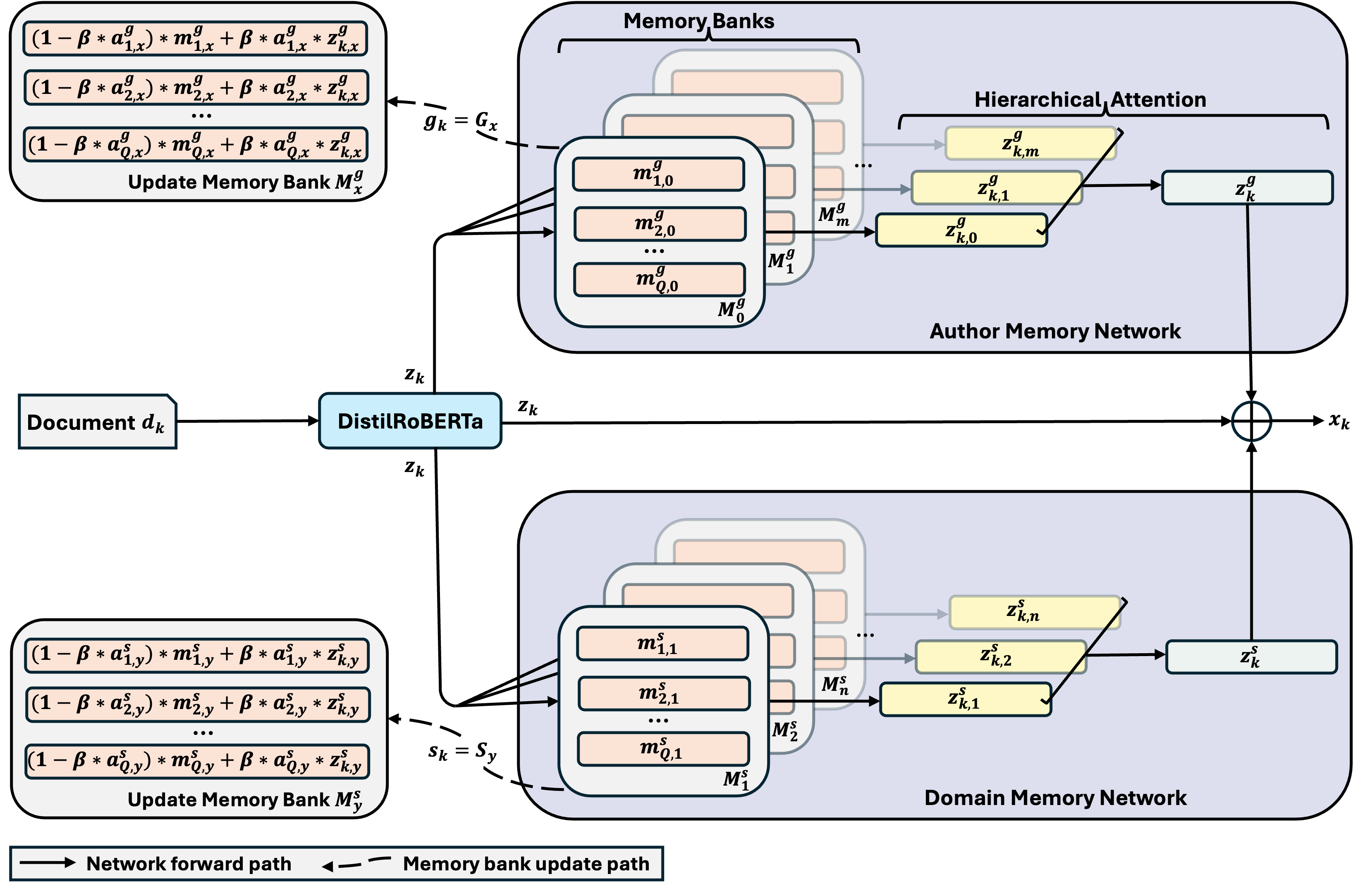}}
{Architecture of Twin Memory Networks (TMN). \label{fig:method-tmn}}
{}
\end{figure}

The architecture of the Twin Memory Networks (TMN) is shown in Figure~\ref{fig:method-tmn}. As depicted, given a training document $d_k$ with its corresponding author label $g_k$ and domain label $s_k$, our design first leverages a pre-trained model (which is DistilRoBERTa \citep{sanh2019distilbert} in this work) to represent the textual content of the document as a $d$-dimensional textual embedding $z_k \in R^d$. 
Next, the TMN utilizes two memory networks with a similar architecture to extract the author and domain embeddings from the input document, respectively. 
In the following subsections, we first detail the design of the author memory network. Next, we describe the domain memory network and illustrate how the final embedding $x_k$ is constructed.

\subsubsection{Author Memory Network}
\noindent 
The goal of the author memory network is to represent the author information for each training document, which we refer to as a document-specific author embedding. To achieve this, we first introduce memory banks to store representations at the author level.
A memory bank is a Neural Turing Machine that has external memory resources to remember its training instances~\citep{graves2014neural}. 
Unlike other embedding methods such as Word2Vec \citep{church2017word2vec} and BERT \citep{devlin2019bert}, memory banks can explicitly store representations and update them through a well-designed mechanism. 
Such memory banks are widely adopted in computer vision, such as video object detection, where they effectively retain representations of past frames \citep{zhou2024rmem}. 

The architecture of our proposed memory banks is shown in Figure~\ref{fig:method-tmn}. Recall from Definition 1, the number of LLMs in the training data is $m$, and the number of authors is $m+1$: humans and $m$ LLMs. 
Each author $G_i$, where $i = 0,1,2 \dots, m$ and $G_0 = H$ (humans), is represented by a memory bank $M^g_i$, where the superscript $g$ indicates that this memory bank is for author representation. Each memory bank contains $Q$ memory units.  
Each memory unit is a $d$-dimensional vector, denoted as $m_{q,i}^g$, where 
$q = 1,2 \dots, Q$, and $i = 0,1,2 \dots, m$.
Memory bank $M^g_i$, which represents author $G_i$, is the concatenation of its component memory units: 
\begin{equation}
\label{eq:representation}
    M^g_i = \left[
        m^g_{1,i} \ m^g_{2,i} \ \cdots \ m^g_{Q,i}
    \right],
\end{equation}
where $i=0,1,2,\dots,m$ and $M^g_i \in R^{d\times Q}$.
Each memory unit $m^g_{q,i}$ represents a specific aspect of author $G_i$, where $q = 1,2 \dots, Q$.
To initialize $m^g_{q,i}$, we denote $Z_i$ as the set of textual embeddings of the training documents authored by $G_i$; that is,  $Z_i=\{z_k ~|~ \forall d_k, g_k = G_i\}$.
We then employ the K-means algorithm to group the textual embeddings in $Z_i$ into $Q$ clusters.
Memory unit $m^g_{q,i}$ is then initialized as the mean embedding of cluster $q$, $q = 1,2 \dots, Q$,
allowing it to capture a specific aspect of author $G_i$. 


Next, we propose a two-level hierarchical attention network to generate a document-specific author embedding for each input document.
As illustrated in Figure~\ref{fig:method-tmn}, the first level of the attention network targets memory units within each memory bank. Specifically, for each memory bank $M^g_i$, $i = 0,1,2 \dots, m$, the attention network assigns higher weights to its memory units that are more relevant to $z_k$ and lower weights to its memory units that are less relevant to $z_k$. Concretely, we have
\begin{equation}
\label{eq:attention}
    \left[ a_{1,i} \ a_{2,i} \ \ldots \ a_{Q,i} \right] = \text{softmax} \left( \frac{z_{k}^T W_a M^g_i}{\tau} \right),    
\end{equation}
where $a_{q,i}$ is the attention weight that measures the relevance of each memory unit $m^g_{q,i}$ in $M^g_i$ to $z_k$, $q = 1,2 \dots, Q$, $[\cdot]^T$ denotes the transpose of a vector or matrix, $W_a \in R^{d\times d}$ is a trainable matrix, and $\tau$ is a temperature hyperparameter.\SingleSpacedXII\footnotemark\footnotetext{The temperature hyperparameter $\tau$ adjusts the smoothness of the attention weight distribution, with a higher value producing a smoother distribution.}\DoubleSpacedXII~
The memory units in $M^g_i$ are then aggregated using their corresponding weights and further transformed by a multi-layer perceptron (denoted as $\text{MLP}_1$) to produce the document-adjusted representation $z_{k,i}^g$ of author $G_i$:
\begin{equation}
\label{eq:attentionweightedsum}
    z^g_{k,i} = \text{MLP}_1 \left(\sum_{q=1}^{Q} a_{q,i} m^g_{q,i} \right), 
\end{equation}
where $z^g_{k,i} \in R^d$ denotes the representation of author $G_i$ adjusted by document $d_k$.

The second level of the attention network takes the document-adjusted author representations $z_{k,i}^g$ as input, $i = 0,1,2 \dots, m$, and assigns weights to them based on their relevance to $z_k$. Concretely, we have:
\begin{equation}
    [b_0 \ b_2 \ \ldots \ b_m]
    = \text{softmax} \left(\frac{z_k^\top W_b [z^g_{k,0} \ z^g_{k,1} \ \ldots \ z^g_{k,m}]
        }{\tau}\right),
\end{equation}
where $b_i$ denotes the attention weight that measures the relevance of $z_{k,i}^g$ to $z_k$, $i = 0,1,2 \dots, m$, $W_b \in R^{d\times d}$ is a trainable matrix, and $\tau$ is the temperature hyperparameter.
Finally, the document-specific author embedding $z_k^g$ for document $d_k$ is calculated as the weighted sum of $z^g_{k,i}$, which is further transformed by a multi-layer perceptron (denoted as $\text{MLP}_2$):
\begin{equation}
    z_k^g = \text{MLP}_2 \left(\sum_{i=0}^{m} b_i z^g_{k,i} \right).
\end{equation}

To better learn the author representations in the memory banks from training data, we propose an updating mechanism. Specifically, given a training document $d_k$ with its author label $g_k=G_x$, where $G_x \in \{G_{0},G_{1},\cdots,G_{m}\}$, the mechanism only updates the memory units in memory bank $M^g_x$, which corresponds to author $G_x$.
 Recall that the first level of the attention network computes the attention weight $a_{q,x}$ for each memory unit $m^g_{q,x}$ in $M^g_x$, as well as the document-adjusted representation $z^g_{k,x}$ of author $G_x$, using Equations \ref{eq:attention} and \ref{eq:attentionweightedsum}. Accordingly, the mechanism updates each memory unit $m^g_{q,x}$ in $M^g_x$ as:
\begin{equation} \label{eq:update}
    m^g_{q,x} = (1 - \beta a_{q,x}) m^g_{q,x} + \beta a_{q,x} z^g_{k,x} ~~~\text{for}
    ~~~q=1, 2, \dots, Q, 
\end{equation}
where $\beta$ is a hyperparameter that adjusts the strength of updating.
Equation~\ref{eq:update} offers three key benefits. First, it uses the author label of a training document to ensure that only the memory units corresponding to that author are updated. This helps learn accurate author representations and prevents unintended updates to other authors' memory units.
Second, after seeing a training document with author label $G_x$, the attention network computes and stores the document-adjusted author information in $z^g_{k,x}$. To reflect this information in the memory units associated with author $G_x$, Equation~\ref{eq:update} updates them  using $z^g_{k,x}$. 
Furthermore, memory units that are more relevant to the document receive larger updates, weighted by $a_{q,x}$.
Third, memory units are fixed during testing. This eliminates the need for author labels during the testing phase, which aligns with the objective of this study, as author labels are unavailable during this phase.

\subsubsection{Domain Memory Network and Summary} 

\noindent 
As shown in Figure~\ref{fig:method-tmn}, the domain memory network has a similar architecture as the author memory network. It aims to extract domain information from each training document and represent it as a document-specific domain embedding.
As illustrated in the figure, each domain $S_i$ is represented by a memory bank $M_i^s$, each of which consists of $Q$ memory units, where $i = 1,2 \dots, n$.
The memory units in $M_i^s$ are initialized by applying the K-means algorithm to the textual embeddings of the training documents belonging to domain $S_i$, where $i = 1,2 \dots, n$. Next, a document-specific domain embedding $z_k^s$ is generated for each training document using a two-level hierarchical attention network similar to that employed in the author memory network, where $k = 1,2 \dots, N$.
The memory units in the domain memory network are then updated using an updating mechanism similar to the one used in the author memory network.
Finally, for each training document $d_k$, its textual embedding $z_k$, document-specific author embedding $z_k^g$, and  document-specific domain embedding $z_k^s$ are concatenated to form a single embedding $x_k$ for the document, where $k = 1,2 \dots, N$.

In summary, we propose a novel TMN design that learns the document-specific author and domain embeddings for each document, in addition to the embedding of its textual content. TMN includes two parallel components: an author memory network and a domain memory network. Each follows a similar architecture, designed to extract the respective author and domain embeddings from a document. 
Take the author memory network as an example. It contains a set of memory banks, each storing information about a specific author. Taking these memory banks as input, TMN extracts the document-specific author embedding from a document using a hierarchical attention network. The memory banks are updated through a novel updating mechanism during training, which selectively determines which memory bank to update and dynamically adjusts the weights of its memory units. 

\subsection{Detection Generalization Module}\label{sec:method-dmg}
Taking the embeddings $x_k$ of training documents as input, for $k = 1,2 \dots, N$, the Detection Generalization Module (DGM) learns a detector that can be generalized to unseen LLMs and domains. 
In the following, we begin with a rigorous theoretical analysis of detection generalization across LLMs and domains. We then present the design of each component of the module, which is guided by the insights from the theoretical analysis.
We denote $\mathcal{D}_{G_i,S_j}=\{x_k|g_k=G_i,s_k=S_j, k=1,2,\dots,N \}$ as the set of embeddings corresponding to training documents generated by LLM $G_i$ in domain $S_j$, where $i=1,2,\dots,m$ and $j=1,2,\dots,n$. Similarly, let $\mathcal{D}_{H,S_j}=\{x_k|g_k=H,s_k=S_j, k=1,2,\dots,N\}$ be the set of embeddings representing training documents written by humans $H$ in domain $S_j$, where $j=1,2,\dots,n$. 
To learn a detector that can discriminate between LLM-generated and human-written information, we construct labeled datasets, each of which contains embeddings corresponding to LLM-generated training documents (labeled $y=1$) and human-written ones (labeled $y=0$). Concretely, each labeled dataset $\mathcal{D}_{ij}=\mathcal{D}_{G_i,S_j}\cup \mathcal{D}_{H,S_j}$ consists of embeddings corresponding to LLM $G_i$-generated and human-written training documents in domain $S_j$, where $i=1,2,\dots,m$ and $j=1,2,\dots,n$.  
We denote the distribution of document embeddings in $\mathcal{D}_{ij}$ as $\mathbb{D}_{ij}$. 
Let $f_{ij}$ be the function that gives the true probability that a document with embedding $x\sim \mathbb{D}_{ij}$ is LLM-generated; that is, $f_{ij} (x)=p(y=1|x)$, where $x\sim \mathbb{D}_{ij}$.
A hypothesis $h$ maps a document with embedding $x$ to its probability of being LLM-generated; i.e., $h(x)=p_h(y=1|x)$.
When applying hypothesis $h$ to document embeddings drawn from the distribution $\mathbb{D}_{ij}$, its error $\epsilon_{ij}(h)$ is defined as the expected difference between $h$ and $f_{ij}$; formally, we have 
\begin{equation}
\label{eq:errorij}
    \epsilon_{ij}(h) = E_{x\sim \mathbb{D}_{ij}} [|h(x)-f_{ij}(x)|],
\end{equation}
where $i=1,2,\dots,m$ and $j=1,2,\dots,n$. 

The objective of the GLID problem is to learn a hypothesis $h$ (i.e., a detector) from the labeled datasets $\bigcup_{\substack{i=1,\dots,m \\ j=1,\dots,n}} \mathcal{D}_{ij}$ such that it generalizes effectively to unseen LLMs and domains. Without loss of generality, let $\mathbb{D}_u$ denote the distribution of embeddings corresponding to documents generated by unseen LLMs or written by humans in unseen domains. By definition, $\mathbb{D}_u$ is unknown. Let $f_{u}$ be the function that gives the true probability that a document with embedding $x\sim \mathbb{D}_u$ is LLM-generated. 
Similarly, we define the error $\epsilon_{u}$ of applying hypothesis $h$ to document embeddings drawn from the distribution $\mathbb{D}_u$ as the expected difference between $h$ and $f_u$:
\begin{equation}
    \epsilon_{u}(h) = E_{x\sim \mathbb{D}_{u}} [|h(x)-f_{u}(x)|].
\end{equation}
Consequently, the objective of the GLID problem becomes learning a hypothesis $h$ from the labeled datasets $\bigcup_{\substack{i=1,\dots,m \\ j=1,\dots,n}} \mathcal{D}_{ij}$ such that $\epsilon_{u}(h)$ is minimized. 
To achieve this, we derive an upper bound for $\epsilon_{u}(h)$ in the following theorem. 

\noindent\textbf{Theorem 1.} \otherlabel{proposition:1}{1} 
\textit{
For any $\eta\in(0,1)$, with probability at least $(1-\eta)^2$, the following bound for $\epsilon_{u}(h)$ holds: }
\begin{equation}
\label{eq:prop1}
\begin{aligned}
\epsilon_u(h) \leq &\sum_{ij} \pi^*_{ij} \epsilon_{ij}(h)+ 
\frac{1}{2} \max_{a,b} \hat{d}_\mathcal{H} (\mathcal{D}_{H,S_a},\mathcal{D}_{H,S_b} ) + \frac{1}{2} \max_{a,b,c,d} \hat{d}_\mathcal{H} (\mathcal{D}_{G_c,S_a},\mathcal{D}_{G_d,S_b} )  + C, \\
\end{aligned}
\end{equation}
\textit{where 
$\pi^*_{ij}\geq 0$, $\sum_{ij}\pi^*_{ij}=1$, $\hat{d}_\mathcal{H}$ is the empirical $\mathcal{H}$-divergence that measures the distance between two distributions using data drawn from the distributions, domain IDs $a,b\in \{1,2,\dots,n\}$, LLM IDs $c,d\in \{1,2,\dots,m\}$,
and $C$ is a small constant. 
}

\textit{\textbf{Proof.} See Appendix~\ref{appendix:proof}. Details of $\pi^*_{ij}$, $\hat{d}_\mathcal{H}$, and $C$ are also given in Appendix~\ref{appendix:proof}.}

By Theorem~\ref{proposition:1}, minimizing the error $\epsilon_u(h)$ requires minimizing all the terms on the right-hand side of Inequality~\ref{eq:prop1}, except for the small constant $C$. 
The first term $\sum_{ij} \pi^*_{ij} \epsilon_{ij}(h)$ is a convex combination of the errors $\epsilon_{ij}(h)$, where $i=1,2,\dots,m$ and $j=1,2,\dots,n$. 
According to Equation \ref{eq:errorij}, $\epsilon_{ij}(h)$ denotes the error of applying hypothesis $h$ to document embeddings drawn from the distribution $\mathbb{D}_{ij}$. Therefore, to minimize the first term, we design a classifier that aims to minimize the error of classifying document embeddings in the labeled dataset $\mathcal{D}_{ij}$ drawn from distribution $\mathbb{D}_{ij}$, for $i=1,2,\dots,m$ and $j=1,2,\dots,n$.

The second term $\max_{a,b} \hat{d}_\mathcal{H} (\mathcal{D}_{H,S_a},\mathcal{D}_{H,S_b})$ is the maximal empirical $\mathcal{H}$-divergence between the distributions of human-written document embeddings in two different domains. To minimize this term, we propose a component that mitigates cross-domain discrepancies among human-written document embeddings. As a result, this component facilitates the learning of domain-invariant embeddings for human-written documents, thereby enabling the effective generalization of our method to unseen domains. 

The third term $\max_{a,b,c,d} \hat{d}_\mathcal{H} (\mathcal{D}_{G_c,S_a},\mathcal{D}_{G_d,S_b})$ represents the maximal empirical $\mathcal{H}$-divergence between the distributions of LLM-generated document embeddings in two different LLM-domain pairs. 
To minimize the third term, we develop a component that reduces both cross-LLM and cross-domain discrepancies among LLM-generated document embeddings, thereby promoting the learning of LLM-invariant and domain-invariant embeddings for LLM-generated documents and allowing our method to generalize effectively to unseen LLMs and domains. 

\begin{figure}[htbp]
\FIGURE{\includegraphics[width=0.95\textwidth]{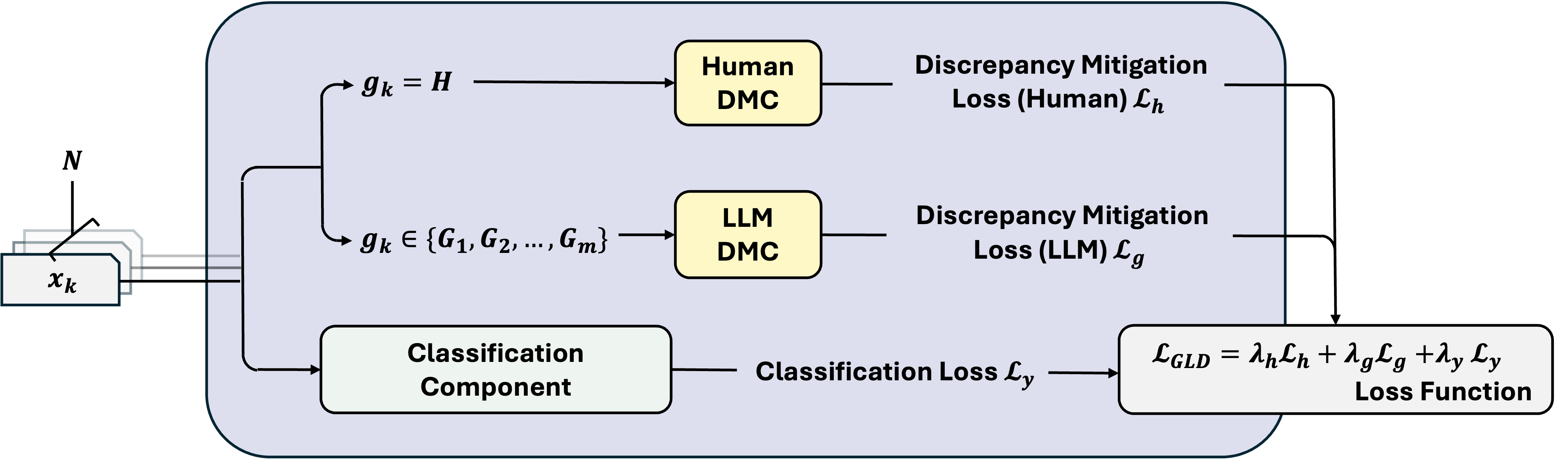}}
{Architecture of Detection
Generalization Module (DGM). \label{fig:method-dmg}}
{}
\end{figure}

In summary, as illustrated in Figure~\ref{fig:method-dmg}, the DGM consists of three components, each of which is designed to minimize a corresponding term in Inequality~\ref{eq:prop1}.  
In the following subsections, we detail the design of each component.

\subsubsection{Discrepancy Mitigation Components}

\noindent The Discrepancy Mitigation Components (DMCs) consist of a Human DMC and an LLM DMC. The former is designed to minimize $\max_{a,b} \hat{d}_\mathcal{H} (\mathcal{D}_{H,S_a},\mathcal{D}_{H,S_b})$, thereby reducing cross-domain discrepancies among human-written document embeddings; the latter aims to minimize $\max_{a,b,c,d} \hat{d}_\mathcal{H} (\mathcal{D}_{G_c,S_a},\mathcal{D}_{G_d,S_b})$, thus mitigating 
cross-LLM and cross-domain discrepancies in LLM-generated document embeddings.
As shown in Figure~\ref{fig:method-dmg}, document embeddings $x_k$ are routed to either the Human DMC or the LLM DMC according to their author labels $g_k$, where $k = 1,2 \dots, N$.
Both the Human DMC and the LLM DMC measure the empirical $\mathcal{H}$-divergence $\hat{d}_\mathcal{H}$ using the widely applied Maximum Mean Discrepancy (MMD) metric with a multi-Gaussian kernel \citep{borgwardt2006integrating, long2015learning}.
Concretely, the MMD between two datasets $\mathcal{D}'$ and $\mathcal{D}''$ is given by:
\begin{equation}
\label{eq:mmd}
    \text{MMD}(\mathcal{D}',\mathcal{D}'') = \frac{1}{N'^2} \sum_{i=1}^{N'} \sum_{j=1}^{N'} \kappa(x_i',x_j') + \frac{1}{N''^2} \sum_{i=1}^{N''} \sum_{j=1}^{N''} \kappa(x_i'',x_j'') - 
    \frac{2}{N'N''} \sum_{i=1}^{N'} \sum_{j=1}^{N''} \kappa(x_i',x_j''), 
\end{equation}
where $N'$ and $N''$ denote the number of instances in $\mathcal{D}'$ and $\mathcal{D}''$, respectively, and $x'$ and $x''$ represent an instance in $\mathcal{D}'$ and $\mathcal{D}''$, respectively. 
The kernel function $\kappa(x, y)$ between two instances $x$ and $y$ is defined as, 
\begin{equation}
\label{eq:kernal}
    \kappa (x,y) = \frac{1}{|R|} \sum_{r \in R} \exp \left( \frac{ - ||x-y||^2}{r} \right),
\end{equation} 
where $||\cdot||$ denotes the $\ell_2$-norm, $r$ is a bandwidth chosen from a bandwidth set $R$ with predefined hyperparameters $r_1$, $r_2$, and a multiplicative step-size of 2, $R =\{2^{r_1},2^{r_1+1},\dots,2^{r_2}\}$, and $|R|$ is the size of the set $R$.

With MMD defined, minimizing $\max_{a,b} \hat{d}_\mathcal{H} (\mathcal{D}_{H,S_a},\mathcal{D}_{H,S_b})$ and $\max_{a,b,c,d} \hat{d}_\mathcal{H} (\mathcal{D}_{G_c,S_a},\mathcal{D}_{G_d,S_b})$ becomes minimizing $\max_{a,b} \text{MMD}  (\mathcal{D}_{H,S_a},\mathcal{D}_{H,S_b})$ and $\max_{a,b,c,d} \text{MMD} (\mathcal{D}_{G_c,S_a},\mathcal{D}_{G_d,S_b})$, respectively.
Therefore, we define the following two losses:
\begin{equation}
\label{eq:loss1}
    \mathcal{L}_{h} = \max_{a,b} \text{MMD}(\mathcal{D}_{H,S_a},\mathcal{D}_{H,S_b} ),
\end{equation}
\begin{equation}
\label{eq:loss2}
    \mathcal{L}_{g} = \max_{\substack{a,b,c,d}} \text{MMD} (\mathcal{D}_{G_c,S_a},\mathcal{D}_{G_d,S_b}). 
\end{equation}
By minimizing the losses $\mathcal{L}_{h}$ and $\mathcal{L}_{g}$, our method learns domain-invariant embeddings for human-written documents and LLM- and domain-invariant embeddings for LLM-generated documents.

\subsubsection{Classification Component and Summary}

\noindent The classification component takes the embedding $x_k$ of each training document as input and outputs the probability $p(\hat{y_k}=1|x_k)$ that the document is LLM-generated, where $k = 1,2 \dots, N$. Specifically, each document embedding is transformed by a multi-layer perceptron (denoted as $\text{MLP}_3$), followed by a sigmoid function. Formally, we have,
\begin{equation}
     p(\hat{y_k}=1|x_k) =  \text{Sigmoid} (\text{MLP}_3 (x_k)),
\end{equation}
where $k = 1,2 \dots, N$. By Theorem~\ref{proposition:1}, the objective of the classification component is to minimize the error in classifying document embeddings. Accordingly, it minimizes the following cross-entropy loss: 
\begin{equation} \label{eq:cls}
\begin{aligned}
    \mathcal{L}_y &= - \sum_{k=1 }^{N} \bigg[y_k \log p(\hat{y_k}=1|x_k) + (1-y_k) \log \big( 1-p(\hat{y_k}=1|x_k) \big) \bigg],
\end{aligned}
\end{equation}
where true document label $y_k$ is 1 if it is LLM-generated or 0 if it is human-written and $k = 1,2 \dots, N$. 

In summary, we propose the DGM module, which learns a detector capable of generalizing to unseen LLMs and domains. The design of the module is guided by a novel theoretical analysis that establishes an upper bound on the generalization error incurred when extending a detector to unseen LLMs and  domains. Specifically, the DGM module consists of three components, each of which is designed to minimize a key term in the upper bound of the generalization error. Together, these components enable our method to effectively generalize to unseen LLMs and domains.


\subsection{Training and Detecting}
\label{sec:method-training}


The loss for the proposed GLD method is the combination of the losses $\mathcal{L}_{h}$, $\mathcal{L}_{g}$, and $\mathcal{L}_{y}$:
\begin{equation}
\label{eq:loss_all}
    \mathcal{L}_{GLD} =  \lambda_h \mathcal{L}_{h} + \lambda_g \mathcal{L}_{g} + \lambda_y \mathcal{L}_y, 
\end{equation}
where $\mathcal{L}_h$, $\mathcal{L}_g$, and $\mathcal{L}_y$ are defined in Equations \ref{eq:loss1}, \ref{eq:loss2}, and \ref{eq:cls}, respectively. 
The non-negative hyperparameters $\lambda_h$, $\lambda_g$, and $\lambda_y$ determine the weights of their corresponding losses. 
As illustrated in Figure~\ref{fig:method-overview}, the GLD method is trained by minimizing its loss $\mathcal{L}_{GLD}$ over the labeled training documents. More specifically, it is trained to learn document embeddings that are both LLM- and domain-invariant by minimizing its component losses $\mathcal{L}_h$ and $\mathcal{L}_g$, and to classify these embeddings effectively by minimizing its component loss $\lambda_y$. As a result, the classification capability of our method is not restricted to documents generated by specific LLMs in particular domains. Instead, it can effectively detect information produced by unseen LLMs in new domains.

Once trained, the GLD method detects whether a test document is LLM-generated or human-written using its two modules, as depicted in Figure~\ref{fig:method-overview}. Specifically, the TMN module produces the embedding of this test document. Since GLD has already been trained, memory units in the TMN module are not updated during the testing phase. Next, the classification component in the DGM module classifies the embedding as LLM-generated or human-written. For the same reason as before, the Human DMC and the
LLM DMC components in this module are not invoked during the testing phase.


\section{EMPIRICAL EVALUATION}
\subsection{Data}
We collected human-written documents across five domains commonly found on digital platforms: news, review, story, general knowledge, and question-and-answer (QA). Specifically, we employed publicly available datasets of human-written documents: the TLDR News dataset\SingleSpacedXII\footnotemark\footnotetext{\url{https://huggingface.co/datasets/JulesBelveze/tldr_news} (last accessed on June 15, 2025)}\DoubleSpacedXII~for news, the Yelp dataset \citep{zhang2015character} for review, the Reddit's ROCStories dataset \citep{mostafazadeh2016corpus} for story, the dataset of Wikipedia articles \citep{Rajpurkar2016SQuAD1Q} for general knowledge, and the ELI5 dataset \citep{fan2019eli5} for QA. For each domain, we randomly sampled 1,500 human-written documents from its corresponding dataset. To produce LLM-generated documents, we utilized five widely adopted commercial or open-source LLMs: GPT-4o\SingleSpacedXII\footnotemark\footnotetext{\url{https://openai.com/index/hello-gpt-4o} (last accessed on June 15, 2025)}\DoubleSpacedXII~from OpenAI, Command-R\SingleSpacedXII\footnotemark\footnotetext{\url{https://cohere.com/blog/command-series-0824} (last accessed on June 15, 2025)}\DoubleSpacedXII~from Cohere
\DoubleSpacedXII, LLaMA-3.1-70B~\citep{dubey2024llama} and OPT-66B \citep{zhang2022opt} from Meta, and GPT-NeoX-20B \citep{black2022gpt} from EleutherAI. For each LLM, we randomly selected 500 human-written documents from each domain. Following the practice of previous studies \citep{mitchell2023detectgpt,bao2023fast}, we then prompted the LLM to produce a LLM-generated document based on the initial 30 words of each selected document.\SingleSpacedXII\footnotemark\footnotetext{We also prompted a LLM to generate text by truncating human-written text to its initial 20 and 40 words and evaluated the results. The outcomes of truncating to 20, 30, and 40 words were similar and showed no statistically significant differences. Therefore, we present only the results for 30 words here.}\DoubleSpacedXII~ This procedure was repeated for each of the five LLMs across all five domains, producing LLM-generated documents for our study. As summarized in Table~\ref{tab:exp-data-statistics}, for each of the five domains, our dataset consists of 1,500 human-written documents and 500 LLM-generated documents per LLM. In total, it contains 20,000 documents.  

\begin{table}[htbp]
\centering
\OneAndAHalfSpacedXI
\caption{Number of Human-written or LLM-generated Documents in Each Domain.}
\label{tab:exp-data-statistics}
\begin{tabular}{p{2.25cm} >{\centering\arraybackslash}p{2cm} >{\centering\arraybackslash}p{2cm} >{\centering\arraybackslash}p{2cm} >{\centering\arraybackslash}p{2cm} >{\centering\arraybackslash}p{2cm} >{\centering\arraybackslash}p{2cm}} \toprule
           & News     & Review    & Story   & Knowledge & QA     & Total  \\ \midrule
Human      & 1,500    & 1,500     & 1,500   & 1,500     & 1,500  & 7,500  \\ \midrule
GPT-4o     & 500      & 500       & 500     & 500       & 500    & 2,500  \\
Command-R  & 500      & 500       & 500     & 500       & 500    & 2,500  \\
LLaMA-3.1  & 500      & 500       & 500     & 500       & 500    & 2,500  \\ 
OPT        & 500      & 500       & 500     & 500       & 500    & 2,500  \\
GPT-NeoX   & 500      & 500       & 500     & 500       & 500    & 2,500  \\ \midrule
Total      & 4,000    & 4,000     & 4,000   & 4,000     & 4,000  & 20,000 \\ 
\bottomrule
\end{tabular}
\end{table}

\subsection{Benchmark Methods and Evaluation Procedure}
\label{sec:procedure}
As reviewed in Section~\ref{sec:rw}, 
current methods for detecting LLM-generated information can be classified into three categories: zero-shot, feature-based, and fine-tuning methods. Therefore, we benchmarked our method against state-of-the-art methods in these three categories. For zero-shot methods, we included DetectGPT \citep{mitchell2023detectgpt}, one of the most widely used techniques based on changes in perplexity before and after perturbing a document to be detected. We also considered Fast-DetectGPT \citep{bao2023fast}, an improvement to DetectGPT, which enhances the original perturbation process through sampling. 
Additionally, we implemented DetectLLM \citep{su2023detectllm}, which differentiates human-written documents from LLM-generated ones based on the average ratio of a token's log-likelihood to its log-rank in a document. 
We chose two representative feature-based methods as benchmarks: GLTR \citep{gehrmann2019gltr} and the multi-feature method \citep{frohling2021feature}. The former trains a classifier to classify a document as human-written or LLM-generated based on probability-based features extracted from the document, while the latter classifies a document based on its linguistic features.
For fine-tuning methods, a commonly used detector is OpenAI's GPT-2 output detector (hereafter referred to as the OpenAI detector) \citep{solaiman2019release}, which identifies GPT-generated information by fine-tuning the RoBERTa model. We also implemented two existing fine-tuning methods as benchmarks: EAGLE \citep{bhattacharjee2024eagle} and DATeD \citep{guo2024ustc}. The former applies adversarial training \citep{ganin2016domain} to enhance the generalizability of a detector to unseen LLMs, while the latter employs adversarial training to enable a detector to generalize to unseen domains. Table~\ref{tab:exp-benchmark-methods} summarizes all the methods compared in our evaluation. Implementation details of each method can be found in Appendix~\ref{appendix:implementation-details}.

\begin{table}[htbp]
\centering
\OneAndAHalfSpacedXI
\caption{Detection Methods Compared in Our Evaluation.}
\def\arraystretch{1.3}
\begin{tabular}{ p{3cm} p{12cm} } 
\toprule 
Method & Notes  \\ 
\midrule
GLD & Our method \\ 
\midrule

DetectGPT & Benchmark zero-shot method \citep{mitchell2023detectgpt} \\
Fast-DetectGPT & Benchmark zero-shot method \citep{bao2023fast} \\
DetectLLM & Benchmark zero-shot method \citep{su2023detectllm} \\ 
\midrule

GLTR & Benchmark feature-based method \citep{gehrmann2019gltr}\\ 
Multi-feature & Benchmark feature-based method \citep{frohling2021feature} \\
\midrule

OpenAI Detector & Benchmark fine-tuning method \citep{solaiman2019release} \\ 
EAGLE & Benchmark fine-tuning method \citep{bhattacharjee2024eagle} \\
DATeD & Benchmark fine-tuning method \citep{guo2024ustc} \\ 

\bottomrule
\end{tabular}%
\label{tab:exp-benchmark-methods}
\end{table}%

We followed the leave-one-group-out (LOGO)  evaluation procedure\SingleSpacedXII\footnotemark\footnotetext{See \url{https://scikit-learn.org/stable/modules/generated/sklearn.model_selection.LeaveOneGroupOut.html} (last accessed on June 15, 2025)}\DoubleSpacedXII~to assess the performance of a detection method on unseen LLMs and domains, in line with the literature \citep{bhattacharjee2024eagle,guo2024ustc}. 
A group in our study consists of human-written documents in a specific domain and documents generated by a specific LLM in the same domain.
An example of this evaluation setup is provided in Table~\ref{tab:exp-leave-one-out}. As shown, to evaluate the performance of a method in detecting information generated by an unseen LLM (i.e., GPT-4o) in an unseen domain (i.e., news), we constructed a training dataset consisting of human-written documents from four other domains (i.e., review, story, knowledge, and QA) and LLM-generated documents produced by four other LLMs (i.e., Command-R, LLaMA-3.1, OPT, and GPT-NeoX) across the four domains. 
The test dataset contained human-written and GPT-4o-generated documents in the news domain. During testing, no information about the tested LLM or domain was provided.
This evaluation setup ensures that learning-based methods (including feature-based and fine-tuning methods) are not exposed to data from the tested LLM or domain, and zero-shot methods have no prior information of the tested LLM or domain. As a result, we can rigorously evaluate the generalizability of each method to unseen LLMs and domains.

\begin{table}[htbp]
\centering
\OneAndAHalfSpacedXI
\caption{An Example of Leave-one-group-out (LOGO) Evaluation for Unseen LLM of GPT-4o and Unseen Domain of News.}
\label{tab:exp-leave-one-out}
\begin{tabular}{p{2.4cm} >{\centering\arraybackslash}p{2.4cm} >{\centering\arraybackslash}p{2.4cm} >{\centering\arraybackslash}p{2.4cm} >{\centering\arraybackslash}p{2.4cm} >{\centering\arraybackslash}p{2.4cm} >{\centering\arraybackslash}p{2.4////cm}} \toprule
           & Review        & Story        & Knowledge    & QA            & News\\ \midrule
Human      & Train (1500)  & Train (1500) & Train (1500) & Train (1500)  & Test (1500)\\ \midrule
Command-R  & Train (500)   & Train (500)  & Train (500)  & Train (500)   & - \\ 
LLaMA-3.1  & Train (500)   & Train (500)  & Train (500)  & Train (500)   & - \\ 
OPT        & Train (500)   & Train (500)  & Train (500)  & Train (500)   & - \\ 
GPT-NeoX   & Train (500)   & Train (500)  & Train (500)  & Train (500)   & - \\ 
GPT-4o     & -             & -            & -            & -             & Test (500) \\ 
\bottomrule
\end{tabular}
\end{table}

To evaluate the performance of each method in Table~\ref{tab:exp-benchmark-methods}, we conducted the LOGO evaluation for all 25 unseen LLM-domain pairs (five LLMs × five domains). 
In each evaluation, we calculated the Area Under the Curve (AUC) and F1 score. Following~\cite{mitchell2023detectgpt}, we reported the AUC to comprehensively evaluate a method's ability to distinguish LLM-generated information from human-written information. Concretely, the AUC is computed by integrating the true positive rate (TPR) with respect to the false positive rate (FPR) over the range $[0, 1]$. We also calculated the F1 score to provide a focused measure of a method's precision and recall in detecting LLM-generated information. In general, let P denote the  number of positive instances, TP be the number of true positives, and FP be the number of false positives. The F1 score is computed as $\text{F1} = \frac{2 \times \text{TP}}{\text{TP} + \text{FP} + \text{P}}$. 



\subsection{Evaluation Results}

Following the procedure stated in Section~\ref{sec:procedure}, we evaluated the performance of each method in Table~\ref{tab:exp-benchmark-methods}. 
Table~\ref{tab:exp-benchmark} reports the average AUC and F1 score for each method across all 25 unseen LLM-domain pairs, with standard deviations provided in parentheses. The evaluation results highlight several findings. 
First, our proposed GLD method surpasses all benchmarks in both AUC and F1 score, and the performance improvements by our method are statistically significant. In particular, our method increases AUC by 21.48\%, 15.18\%, and 5.71\% and lifts F1 score by 25.72\%, 20.41\%, and 7.13\% over the best performing zero-shot, feature-based, and fine-tuning benchmarks, respectively. 
Second, our method achieves the lowest standard deviations in both AUC and F1 score, among all the compared methods. This indicates that our method demonstrates greater stability across different combinations of unseen LLM and domain, exhibiting less performance bias toward specific combinations. 
Third, among the benchmarks, fine-tuning methods outperform both feature-based and zero-shot methods, which supports our choice of a fine-tuning approach for developing our method.
In summary, the evaluation results demonstrate the greater effectiveness and stability of our method in distinguishing human-written information from content generated by unseen LLMs in new domains, compared to state-of-the-art benchmark methods. The superior performance of our method can be attributed to its key innovations: the twin memory networks for representing the content, author, and domain of each piece of information, and the detection generalization module for learning LLM- and domain-invariant embeddings.



\begin{table}[htbp]
\centering
\OneAndAHalfSpacedXI
\caption{Performance Comparison between GLD and Benchmark Methods.}
\def\arraystretch{1.3}
\begin{tabular}{ p{3cm} p{2.5cm} p{2.5cm} p{2.5cm} p{2.5cm} }
\toprule
Method & AUC & Improvement by GLD & F1 score & Improvement by GLD \\
\midrule
\multirow{2}{*}{GLD} & 0.888 & & 0.826 & \\ \addlinespace[-1ex]
& (0.085) & & (0.098) \\
\midrule

\multirow{2}{*}{DetectGPT} & 0.561$^{**}$ & 58.29\% & 0.438$^{**}$ & 88.58\% \\ \addlinespace[-1ex]
& (0.115) &  & (0.132) \\
\multirow{2}{*}{Fast-DetectGPT} & 0.632$^{**}$ & 40.51\% & 0.572$^{**}$ & 44.41\% \\ \addlinespace[-1ex]
& (0.363) &  & (0.368) \\
\multirow{2}{*}{DetectLLM} & 0.731$^{**}$ & 21.48\% & 0.657$^{**}$ & 25.72\% \\ \addlinespace[-1ex]
& (0.133) &  & (0.157) \\
\midrule

\multirow{2}{*}{GLTR} & 0.628$^{**}$ & 41.40\% & 0.543$^{**}$ & 52.12\% \\ \addlinespace[-1ex]
& (0.197) & & (0.201) \\
\multirow{2}{*}{Multi-feature} & 0.771$^{**}$ & 15.18\% & 0.686$^{**}$ & 20.41\% \\ \addlinespace[-1ex]
& (0.138) & & (0.162) \\
\midrule

\multirow{2}{*}{OpenAI Detector} & 0.834$^{**}$ & 6.47\% & 0.757$^{**}$ & 9.11\% \\ \addlinespace[-1ex]
& (0.200) & & (0.235) \\
\multirow{2}{*}{EAGLE} & 0.840$^{**}$ & 5.71\% & 0.771$^{**}$ & 7.13\% \\ \addlinespace[-1ex]
& (0.147) & & (0.169) \\
\multirow{2}{*}{DATeD}  & 0.821$^{**}$ & 8.16\% & 0.740$^{**}$ & 11.62\% \\ \addlinespace[-1ex]
& (0.167) & & (0.256) \\

\bottomrule 
\end{tabular}%
\begin{tablenotes}
\centering\item[*] Note: The notation ** indicates that the performance difference between GLD and a benchmark is statistically significant at the 5\% level. Standard deviations are provided in parentheses.
\end{tablenotes}
\label{tab:exp-benchmark}
\end{table}



\begin{figure}[htbp]
\FIGURE{\includegraphics[width=0.95\textwidth]{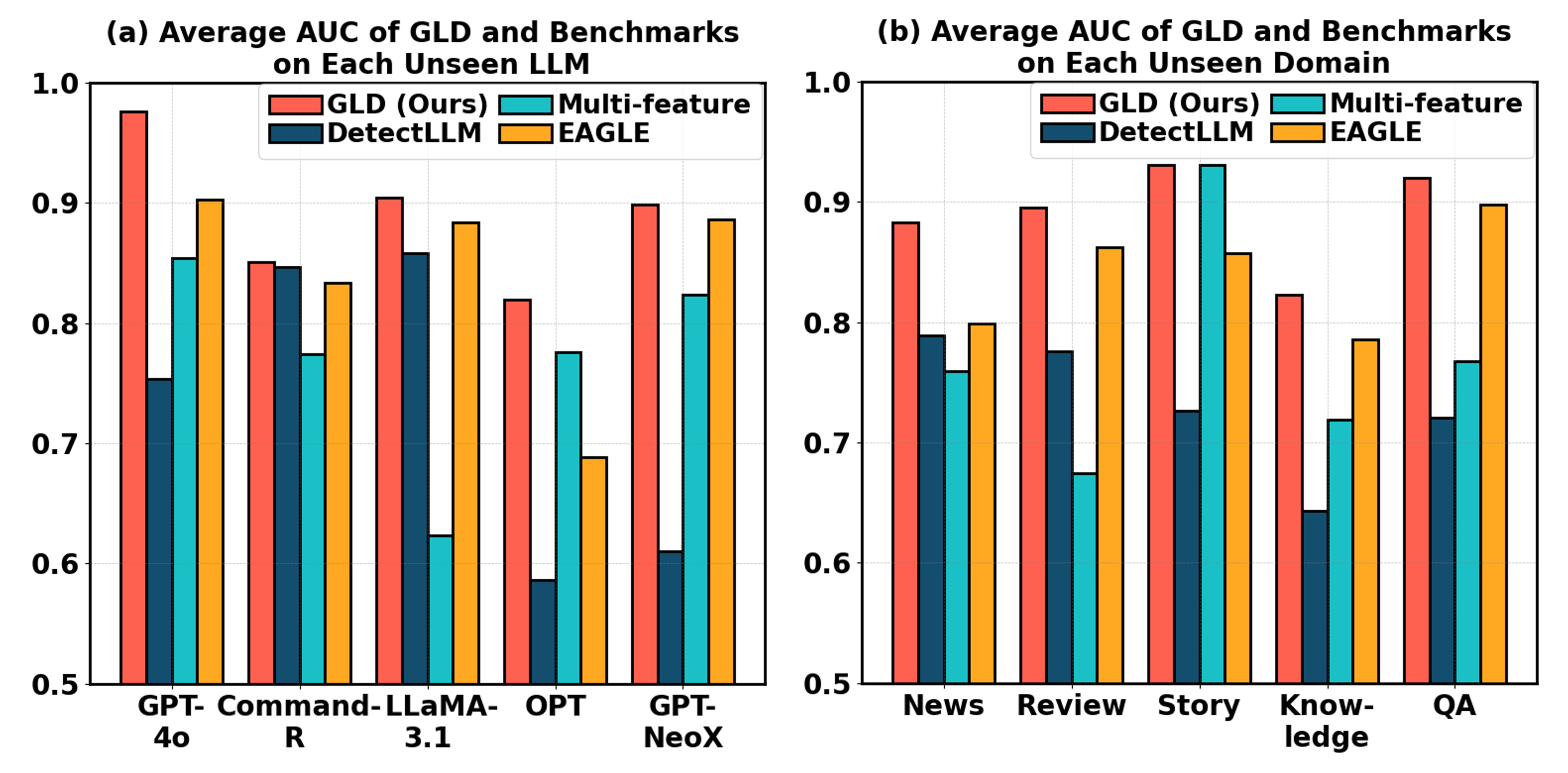}}
{Average AUC of GLD and Best Performing Benchmark Methods on Each Unseen LLM or Domain. \label{fig:exp-avg-LLM-domain}}
{Better view in color. }
\end{figure}

To provide a deeper insight into the performance of our proposed GLD method, we present its performance on each unseen LLM, alongside that of the best performing benchmark in each category: DetectLLM for zero-shot methods, the multi-feature method for feature-based methods, and EAGLE for fine-tuning methods. As reported in Figure~\ref{fig:exp-avg-LLM-domain}(a), 
our method outperforms all benchmarks in AUC on each unseen LLM. 
Furthermore, our method exhibits greater stability across different unseen LLMs. 
Taking DetectLLM as an example, its average AUC of detecting information generated by an unseen LLM ranges from 0.586 (when detecting information generated by OPT) to 0.858 (when detecting information generated by LLaMA-3.1). 
In contrast, our method consistently achieves an AUC greater than 0.8 across all unseen LLMs.
Figure~\ref{fig:exp-avg-LLM-domain}(b) presents the performance of our method and the top performing benchmarks on each unseen domain. 
As depicted, our method surpasses all benchmarks in AUC on each unseen domain, while demonstrating greater stability across these domains. For example, although the performance of the multi-feature method is close to that of our method in the story domain, the benchmark's performance drops significantly in other domains, falling well behind our method in these domains.
In summary, our additional analysis further confirms the effectiveness and stability of our method in detecting information generated by various unseen LLMs across different unseen domains.
As a result, our method is particularly useful for identifying LLM-generated information, given that LLMs are increasingly being applied to a growing range of domains and new LLMs continue to emerge rapidly.

\subsection{Performance Analysis}

We conducted ablation studies to assess the contribution of each key design artifact of GLD to its performance. 
In general, we dropped one artifact from GLD at a time. The performance difference between the resulting method without the dropped artifact and GLD revealed the performance contribution of the artifact. 
As illustrated in Figure~\ref{fig:method-overview}, the GLD method consists of two key modules: TMN and DGM. To evaluate the performance contribution of each module, we conducted two sets of ablation studies.
In the first set of studies, we removed the TMN module entirely from GLD, referring to the resulting variant as GLD w/o TMN.
Furthermore, to evaluate the contribution of each memory network in TMN, we removed either the author memory network or the domain memory network individually. The resulting variants were referred to as GLD w/o Author MN and GLD w/o Domain MN, respectively. The core innovations of the DGM module lie in its two Discrepancy Mitigation Components (DMCs). Accordingly, in the second set of studies, we removed both DMCs together or each one individually. The resulting variants were denoted as GLD w/o DMC, GLD w/o Human DMC, and GLD w/o LLM DMC, respectively.

\begin{table}[htbp]
\centering
\OneAndAHalfSpacedXI
\caption{Ablation Study of GLD.}
\def\arraystretch{1.3}
\begin{tabular}{ p{4.7cm} p{2.2cm} p{2.5cm} p{2.2cm} p{2.5cm} }
\toprule 
Method & AUC & Improvement by GLD & F1-score & Improvement by GLD \\
\midrule
\multirow{2}{*}{GLD} & 0.888 & & 0.826 & \\ \addlinespace[-1ex]
& (0.085) & & (0.098) \\ \midrule
\multirow{2}{*}{GLD w/o TMN} & 0.861$^{**}$ & 3.14\% & 0.799$^{**}$ & 3.38\% \\ \addlinespace[-1ex]
& (0.100) & & (0.109) \\
\multirow{2}{*}{GLD w/o Author MN} & 0.870$^{**}$ & 2.07\% & 0.809$^{**}$ & 2.10\% \\ \addlinespace[-1ex]
& (0.107) & & (0.132) \\
\multirow{2}{*}{GLD w/o Domain MN} & 0.869$^{**}$ & 2.19\% & 0.807$^{**}$ & 2.35\% \\ \addlinespace[-1ex]
& (0.111) & & (0.135) \\ \midrule
\multirow{2}{*}{GLD w/o DMC} & 0.858$^{**}$ & 3.38\% & 0.796$^{**}$ & 3.64\% \\ \addlinespace[-1ex]
& (0.113) & & (0.142) \\
\multirow{2}{*}{GLD w/o Human DMC} & 0.863$^{**}$ & 2.90\% & 0.802$^{**}$ & 2.99\% \\ \addlinespace[-1ex]
& (0.121) & & (0.148) \\
\multirow{2}{*}{GLD w/o LLM DMC} & 0.867$^{**}$ & 2.42\% & 0.805$^{**}$ & 2.23\% \\ \addlinespace[-1ex]
& (0.105) & & (0.128) \\
\bottomrule 
\end{tabular}%
\begin{tablenotes}
\centering\item[*] Note: The notation ** indicates that the performance difference between GLD and a benchmark is statistically significant at the 5\% level. Standard deviations are provided in parentheses.
\end{tablenotes}
\label{tab:exp-ablation}
\end{table}

Using the evaluation procedures described in Section~\ref{sec:procedure}, we compared the performance of GLD with its ablated variants.
Table~\ref{tab:exp-ablation} reports the average performance of these methods across 25 unseen LLM-domain pairs. As reported, GLD significantly outperforms all ablated variants, demonstrating the contribution of each novel design artifact to its performance. For example, GLD surpasses GLD w/o TMN by $3.14\%$ in AUC and $3.38\%$ in F1-score, revealing the contribution of the TMN module to its performance.
Diving into different components in TMN, removing the author MN or domain MN separately yields better performance than removing the entire TMN, highlighting the importance of constructing memory networks for both author and domain. Similarly, removing the human DMC or LLM DMC separately also performs better than removing both, corroborating the effectiveness of each component.


\subsection{Case Studies}
\begin{figure}[htbp]
\FIGURE{\includegraphics[width=\textwidth]{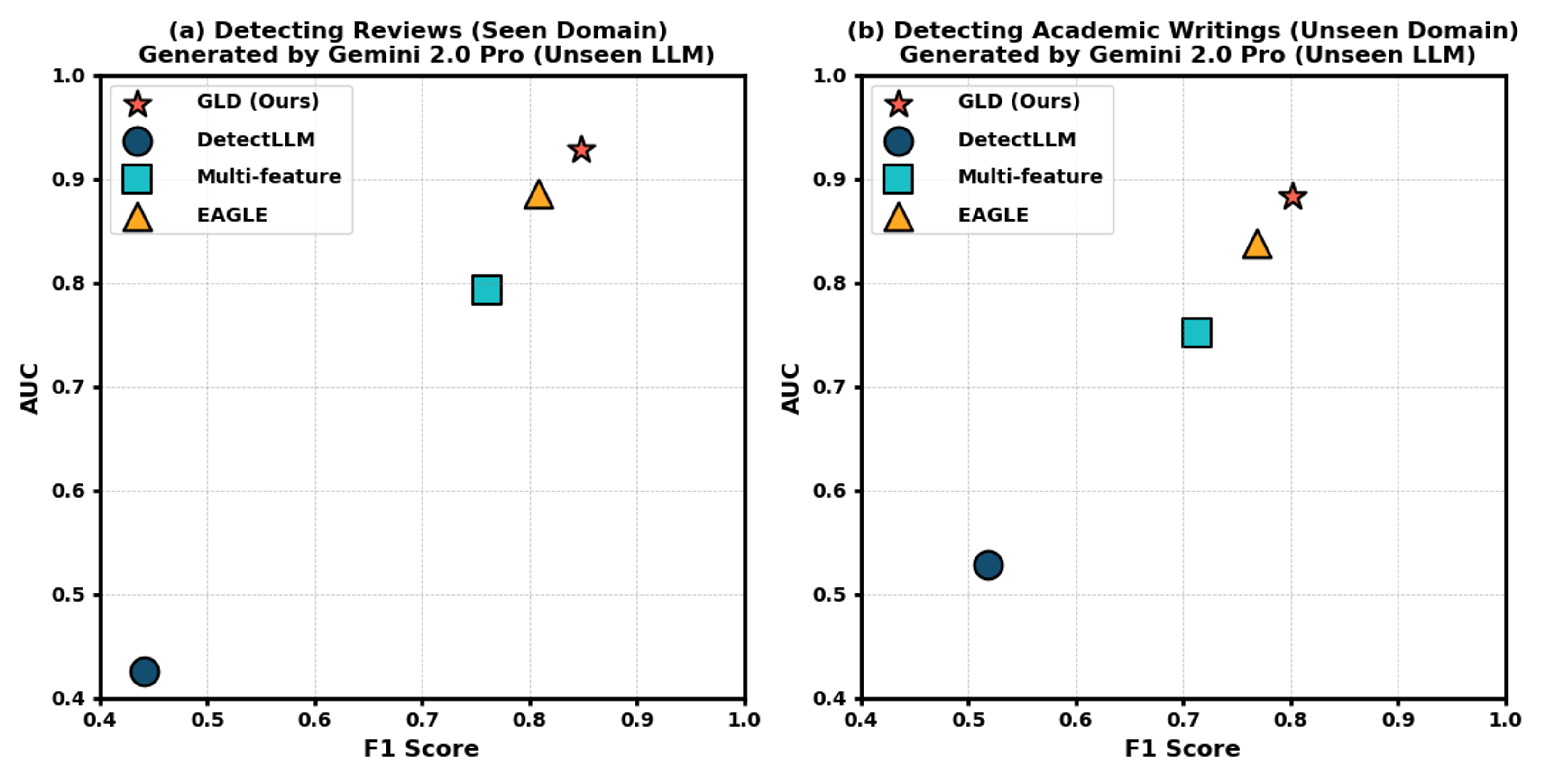}}
{Case Studies on Detecting Reviews and Academic Writings Generated by Gemini 2.0 Pro. \label{fig:exp-case}}
{}
\end{figure}

We conducted two case studies to evaluate the effectiveness of our proposed GLD method in real-world scenarios. In the first case study, we focused on detecting LLM-generated reviews, which is critical for e-commerce platforms and review websites such as Amazon and Yelp to enhance product evaluation and assist consumers in making informed decisions \citep{lappas2016impact}. It was recently shown that an increasing number of fake reviews have been generated by LLMs.\SingleSpacedXII\footnotemark\footnotetext{See \url{https://originality.ai/blog/g2-reviews-ai-generated-case-study} (last accessed on June 15, 2025)}\DoubleSpacedXII~A major challenge in detecting these generated reviews is the rapid emergence of new LLMs, which makes it infeasible to collect sample reviews created by all such models as training data \citep{gupta2024recent}. To address this, we evaluated our method’s ability to detect reviews generated by Gemini 2.0 Pro, an LLM recently developed by Google.\SingleSpacedXII\footnotemark\footnotetext{See \url{https://blog.google/technology/google-deepmind/gemini-model-updates-february-2025} (last accessed on June 15, 2025)}\DoubleSpacedXII~Specifically, we compared the performance of our method against that of the top performing benchmarks in each category: DetectLLM for zero-shot methods, the multi-feature method for feature-based methods, and EAGLE for fine-tuning methods. 
Each method was trained using the complete dataset in Table~\ref{tab:exp-data-statistics}. To construct the test dataset to evaluate the performance of each method, we collected 1,000 human-written Yelp reviews distinct from those in the training dataset and used Gemini 2.0 Pro to generate corresponding fake reviews by continuing from the first 30 words\SingleSpacedXII\footnotemark\footnotetext{Same as in the main experiment, we evaluate the generated text by truncating the human-written text to 20 and 40 words. Since the results have no statistical difference, we report only the 30-word results here.}\DoubleSpacedXII~of each original review. This setup allowed us to evaluate each method’s ability to detect reviews generated by an unseen LLM (i.e., Gemini 2.0 Pro) in a seen domain (i.e., review). The evaluation results presented in Figure~\ref{fig:exp-case}(a) show that our method outperforms all top performing benchmarks in both AUC and F1 score. 
The performance advantage of our method can be attributed to its methodological novelties, which enable it to effectively generalize to unseen LLMs, such as Gemini 2.0 Pro in this case. As a result, our method is particularly useful for detecting LLM-generated reviews and supporting customers on e-commerce platforms, especially in light of the rapid emergence of new LLMs and the growing trend of using LLMs to generate reviews. 

We conducted a second case study to evaluate the performance of our method and the top performing benchmarks in detecting academic writings generated by Gemini 2.0 Pro. 
LLMs have been considered a double-edge sword in academic work due to their ability to assist it, such as improving grammar, while also raising concerns about academic integrity when misused, such as generating an entire section or even a full paper without disclosure. 
Previous research has shown that even human scientists have difficulties distinguishing LLM-generated abstracts from human-written ones \citep{else2023abstracts}. Therefore, it is essential to develop a method that can accurately detect LLM-generated academic writings.
Specifically, we focused on the medical domain. Using the widely recognized PubMed dataset \citep{jin2019pubmedqa}, we collected 1,000 titles and abstracts, along with their conclusions, from research papers authored by human experts in the medical field. We then prompted Gemini 2.0 Pro with each pair of collected title and abstract to generate the corresponding conclusion, in line with the common practice in the literature \citep{jin2019pubmedqa, gilson2023does}. Our test dataset consists of 1,000 human-written conclusions and 1,000 conclusions generated by Gemini 2.0 Pro.
Each method was trained using the complete dataset in Table~\ref{tab:exp-data-statistics} and evaluated using the test dataset.
As a result, both the LLM (i.e., Gemini 2.0 Pro) and the domain (i.e., academic writing) are unseen by any of the methods.
As reported in Figure~\ref{fig:exp-case}(b), our method outperforms all benchmarks in both AUC and F1 score. 
With the continued rollout of new LLMs and their applications to an increasing number of domains, our method is particularly valuable for its ability to generalize effectively to unseen LLMs and domains.

\section{DISCUSSION}

\subsection{Summary and Contributions}

Detecting LLM-generated information is essential for preserving trust on digital platforms and preventing the spread of misinformation. Existing detection methods focus on identifying content generated by specific LLMs in known domains but face significant challenges in generalizing to a large number of unseen LLMs and a wide range of new domains. To address these challenges, we propose the GLD method, which is capable of detecting information produced by unseen LLMs in new domains. Its capabilities are realized through the twin memory networks design and the theory-guided detection generalization module. Extensive empirical evaluations with real-world datasets demonstrate the superiority of our method over state-of-the-art detection methods in identifying information generated by unseen LLMs in new domains.

Our research contributes to the extant IS literature in two ways. First, our work falls into the computational genre of design science research \citep{rai2017editor,padmanabhan2022machine}, which develops novel computational methods to solve important business and societal problems and makes methodological contributions to the literature \citep{samtani2022linking,zhao2023exploiting,kim2023rolex}. 
Driven by the need to detect information generated by unseen LLMs in unseen domains, we propose a detection method that features two methodological contributions: the novel twin memory networks design and the innovative  detection generalization module. Second, an active stream of IS research focuses on the detection and management of fake and misleading content on digital platforms \citep{abbasi2010detecting,moravec2019fake,wei2022combining}. As LLMs are increasingly used to generate such content due to their low costs and high capabilities, our study contributes to this IS research stream by proposing a novel method that can effectively detect information generated by unseen LLMs in new domains. 
Moreover, our work also adds to the growing body of design science research on LLMs in the IS discipline \citep[e.g.,][]{wang2024learning, zhang2024ketch} and advances our knowledge of LLM-based design.




\subsection{Practical Implications}

Our study offers practical implications for digital platforms. The rapid proliferation of LLMs has made it easy for individuals to generate and spread large amounts of fake, misleading, or useless content on digital platforms, undermining the integrity of these platforms \citep{he2022market,jakesch2023human}. Therefore, detecting LLM-generated content is critical for protecting the integrity of digital platforms.
Our study provides a detection method that is trained on a dataset of human-written content and information generated by a small number of LLMs in several domains, but can be generalized to detect information produced by unseen LLMs in new domains. 
Given the vast and growing number of available LLMs and the wide range of domains on digital platforms, our method is especially effective for detecting and managing LLM-generated content on these platforms.
Specifically, digital platform operators can apply our GLD method to detect LLM-generated content and take appropriate actions, such as flagging or removing it. For example, once our method identifies LLM-generated content, community notes could be added to alert users and caution them against potential misinformation.
Moreover, empowered by our method, digital platform operators can monitor accounts that frequently disseminate large volumes of LLM-generated content, allowing them to regulate or even block such accounts \citep{lee2024disinformation}. In summary, our method enables digital platform operators to effectively identify and manage LLM-generated content on the platform, thereby preserving user trust in the platform and safeguarding its business value amid the growing prevalence of LLMs. 

Our study also supports the enforcement of government regulations on LLMs, particularly those aimed at ensuring their accountability \citep{stokel2023chatgpt}. 
Notable among these regulations is the European Union's Artificial Intelligence Act, which mandates that content generated by LLMs must be clearly labeled. Another key regulation is Executive Order 14110, issued by the U.S. government, which directs federal agencies to ensure the responsible deployment of AI systems, including LLMs. Understandably, identifying LLM-generated content is essential for ensuring accountability. To this end, our study offers an efficient and effective tool for detecting and labeling LLM-generated content.
Moreover, government agencies could leverage our GLD method to trace and regulate individuals who frequently publish LLM-generated content without clear labeling. In particular, malicious authors who regularly publish misleading LLM-generated content could be exposed and banned from publishing, thus reducing the harm caused by the misuse of LLMs \citep{pennycook2021psychology}.



\subsection{Future Work}
Our study can be extended in several directions. First, our method focuses on detecting text generated by LLMs. Future research could expand this approach to multimodal settings, such as detecting AI-generated content that combines text with images or audio. This would require generalizing the GLD method to encompass AI systems beyond LLMs. 
Second, our empirical evaluations are based on documents generated by five representative LLMs in five important domains. Future work could further assess the performance of GLD in detecting text generated by other prominent LLMs (e.g., Claude and DeepSeek) in additional domains (e.g., science and finance).
Third, existing literature has explored injecting watermarks in LLMs \citep{kirchenbauer2023watermark}. These watermarking methods typically require a white-box LLM in order to modify its text generation. Although our method cannot be directly applied or guide the design of watermarking methods, future work can investigate whether our GLD method is capable of detecting texts generated by watermarked LLMs. 
Fourth, our study frames the problem as a binary classification task, distinguishing LLM-generated content from human-written text. As the adoption of LLMs becomes more widespread, humans are increasingly using these models to assist with tasks such as proofreading or enhancing their writing. This differs from malicious applications, such as generating fake or misleading content. Future studies should investigate how to detect these diverse scenarios of LLM usage and identify cases where harmful intent is present.
\bibliographystyle{apalike}
\renewcommand*{\bibfont}{\normalsize}
\bibliography{literature}


\newpage

\begin{appendices}

\setcounter{figure}{0}
\setcounter{table}{0}
\setcounter{section}{0}
\setcounter{equation}{0}
\setcounter{page}{1}

\renewcommand{\thetable}{A\arabic{table}}
\renewcommand{\thefigure}{A\arabic{figure}}
\renewcommand{\theequation}{A\arabic{equation}}
\renewcommand{\thepage}{EC\arabic{page}}

\titleformat{\section}{\centering\normalfont\rmfamily\normalsize\bfseries}  {APPENDIX \thesection}{1em}{}

\titleformat{\subsection}{\centering\normalfont\rmfamily\normalsize\bfseries}  {Appendix \thesubsection}{1em}{}

\section{PROOFS OF THEOREM 1}
\label{appendix:proof}


We first introduce the definition of $\mathcal{H}$-divergence \citepec{ben2010theory2}, a widely used metric for measuring the distance between two distributions, as well as the definition of its empirical estimation.

\vspace{1em}
\noindent\textbf{Definition A.1. ($\mathcal{H}$-divergence).} \textit{Let $\mathcal{H}$ denote a hypothesis space, and let $\mathbb{D}'$ and $\mathbb{D}''$ be two distributions. The identity set of a hypothesis $h\in\mathcal{H}$ is $I(h)$; that is, $x\in I(h) \iff h(x)=1$. The $\mathcal{H}$-divergence between $\mathbb{D}'$ and $\mathbb{D}''$ is defined as \citepec{ben2010theory2}}
\begin{equation}
\label{eq:hdivergence}
    d_{\mathcal{H}} (\mathbb{D}', \mathbb{D}'') = 2 \sup_{h\in \mathcal{H}} |P[I(h)|x\sim \mathbb{D}']- P[I(h)|x\sim \mathbb{D}'']|.
\end{equation}

\vspace{1em}
\noindent\textbf{Definition A.2. (Empirical $\mathcal{H}$-divergence).} \textit{Let $\mathcal{D}'$ and $\mathcal{D}''$ be two datasets drawn from distributions $\mathbb{D}'$ and $\mathbb{D}''$, respectively. The $\mathcal{H}$-divergence between the two distributions can be empirically estimated using the two datasets, and the estimation is given by \citepec{ben2010theory2}}
\begin{equation} \label{eq:em_hdiver}
    \hat{d}_{\mathcal{H}}\left(\mathcal{D}', \mathcal{D}''\right)=
    2 \left(1-\min _{h \in \mathcal{H}}\left[\frac{1}{N} \sum_{x: h(x) = 0 } I[x \in \mathcal{D}'] + \frac{1}{N} \sum_{x: h(x) =1} I\left[x \in \mathcal{D}'' \right]\right]\right),
\end{equation}
\textit{where the hypothesis space $\mathcal{H}$ is symmetric (i.e., if $h\in\mathcal{H}~\text{then}~1-h\in\mathcal{H}$), indicator function $I[x \in \mathcal{D}']$ is 1 if $x$ belongs to the dataset $\mathcal{D}'$, and $N$ is the number of instances in each dataset. }

We introduce the following lemmas, which will be used to prove the theorem. Following \citetec{wang2022generalizing2}, we restate Theorem 1 from \citetec{albuquerque2019generalizing2} in the context of our study as Lemma 1. 

\vspace{1em}
\noindent\textbf{Lemma 1.} \otherlabel{lemma:1}{1} \textit{Let $\Lambda$ be the convex hull constructed from document embedding distributions $\mathbb{D}_{ij}$, $i=1,2,\dots,m$ and $j=1,2,\dots,n$; that is, $\Lambda = \sum_{ij} \pi_{ij} \mathbb{D}_{ij}$, where $\pi_{ij}\geq 0$ and $\sum_{ij}\pi_{ij}=1$. Let $\gamma=\min_{\pi  } d_{\mathcal{H}} (\mathbb{D}_{u}, \sum_{ij} \pi_{ij} \mathbb{D}_{ij})$, where $\pi = \{\pi_{ij} | i=1,2,\dots,m, j=1,2,\dots,n\}$; 
the optimal is achieved when $\pi=\pi^*$. Let $\rho=\sup_{i_1,j_1,i_2,j_2} d_\mathcal{H} (\mathbb{D}_{i_1j_1},\mathbb{D}_{i_2j_2})$ be the diameter of the convex hull, where $i_1,i_2\in \{1,2,\dots,m\}$ and $j_1,j_2\in \{1,2,\dots,n\}$. 
The following bound for $\epsilon_u(h)$ holds: }
\begin{equation}
\label{eq:lemma1}
\begin{aligned}
\epsilon_u(h) \leq &\sum_{ij} \pi^*_{ij} \epsilon_{ij}(h)+ 
\frac{\gamma + \rho}{2}  + \epsilon^{*} \\
\end{aligned}
\end{equation}
\textit{where $\pi^* = \{\pi^*_{ij} | i=1,2,\dots,m, j=1,2,\dots,n\}$ and $\epsilon^{*}$ denotes the ideal joint risk, which is the minimum error achieved by any hypothesis in $\mathcal{H}$ over the unseen document embedding distribution $\mathbb{D}_u$ and the distribution $\mathbb{D}^* = \sum_{ij} \pi_{ij}^* \mathbb{D}_{ij}$.}

\textit{\textbf{Proof.} See \citetec{albuquerque2019generalizing2}.  
}

Lemma~\ref{lemma:1} provides an upper bound on the error $\epsilon_u(h)$ incurred when classifying documents drawn from the unseen distribution $\mathbb{D}_u$. 
Note that both $\gamma$ and $\epsilon^{*}$ attain their respective minimum values. Therefore, to reduce $\epsilon_u(h)$, we 
need to reduce the errors $\epsilon_{ij}(h)$, $i=1,2,\dots,m, j=1,2,\dots,n$, and the diameter $\rho$ of the convex hull, the upper bound of which is given by the following lemma.

\vspace{1em}
\noindent\textbf{Lemma 2.} \otherlabel{lemma:2}{2} \textit{
The following bound for the diameter $\rho$ of the convex hull holds: }
\begin{equation}\label{eq:lemma2}
\begin{aligned}
\rho & \leq  \max_{a,b} {d}_\mathcal{H} (\mathbb{D}_{H,S_a},\mathbb{D}_{H,S_b} ) + \max_{a,b,c,d} {d}_\mathcal{H} (\mathbb{D}_{G_c,S_a},\mathbb{D}_{G_d,S_b} ) + 4 \max_{\alpha_{i_1j_1},\alpha_{i_2j_2}} \left| \alpha_{i_1j_1} - \alpha_{i_2j_2} \right|, 
\end{aligned}
\end{equation}
\textit{where domain IDs $a, b \in \{1,2,\dots,n\}$, LLM IDs $c, d\in \{1,2,\dots,m\}$, and $\alpha_{i_1j_1}$ is the proportion of human-written documents in the documents drawn form distribution $\mathbb{D}_{i_1j_1}$, $i_1=1,2,\dots,m$ and $j_1=1,2,\dots,n$. 
}

\textit{\textbf{Proof.} By definition, $\rho=\sup_{i_1,j_1,i_2,j_2} d_\mathcal{H} (\mathbb{D}_{i_1j_1},\mathbb{D}_{i_2j_2})$. 
Without loss of generality, let the supremum achieved when $i_1=i_1^*$, $j_1=j_1^*$, $i_2=i_2^*$, and $j_2=j_2^*$. That is, $\rho= d_\mathcal{H} (\mathbb{D}_{i_1^*j_1^*},\mathbb{D}_{i_2^*j_2^*})$}. For simplicity of notation, we denote $P(h(\mathbb{D}))=P[I(h)|x\sim \mathbb{D}]$.
\begin{equation}
\begin{aligned}
\rho & = d_\mathcal{H} (\mathbb{D}_{i_1^*j_1^*},\mathbb{D}_{i_2^*j_2^*}) \\
& = 2 \sup_{h\in\mathcal{H}} | P [h(\mathbb{D}_{i_1^*j_1^*})]- P [h(\mathbb{D}_{i_2^*j_2^*})] | \\ 
& = 2 \sup_{h\in\mathcal{H}} \left| 
\alpha_{i_1^*j_1^*} P [h(\mathbb{D}_{H,S_{j_1^*}})] + 
(1-\alpha_{i_1^*j_1^*}) P [h(\mathbb{D}_{G_{i_1^*},S_{j_1^*}})] \right. \\ 
& ~~~~~~~~~~~~~~~ - \left. \alpha_{i_2^*j_2^*} P [h(\mathbb{D}_{H,S_{j_2^*}})] - (1-\alpha_{i_2^*j_2^*}) P [h(\mathbb{D}_{G_{i_2^*},S_{j_2^*}})] \right| \\
& \leq 2 \sup_{h\in\mathcal{H}} \bigg\{ \left| 
\alpha_{i_1^*j_1^*} P [h(\mathbb{D}_{H,S_{j_1^*}})]  -
\alpha_{i_2^*j_2^*} P [h(\mathbb{D}_{H,S_{j_2^*}})] \right|  \\ 
& ~~~~~~~~~~~~~~~ + \left|(1-\alpha_{i_1^*j_1^*}) P [h(\mathbb{D}_{G_{i_1^*},S_{j_1^*}})]  -
(1-\alpha_{i_2^*j_2^*}) P [h(\mathbb{D}_{G_{i_2^*},S_{j_2^*}})]  \right| \bigg\}. \\
\end{aligned}
\end{equation}
\textit{The last step in (A5) follows from the triangle inequality for the absolute function, i.e., $|a+b|\leq|a|+|b|$. By adding and subtracting $\alpha_{i_1^*j_1^*} P [h(\mathbb{D}_{H,S_{j_2^*}})]$ in the first absolute term, and adding and subtracting $\alpha_{i_1^*j_1^*} P [h(\mathbb{D}_{G_{i_2^*},S_{j_2^*}})]$ in the second absolute term, we obstain the following inequality. }

\begin{equation}
\begin{aligned}
\rho 
& \leq 2 \sup_{h\in\mathcal{H}} \bigg\{ \left| 
\alpha_{i_1^*j_1^*} P [h(\mathbb{D}_{H,S_{j_1^*}})]  -
\alpha_{i_1^*j_1^*} P [h(\mathbb{D}_{H,S_{j_2^*}})] + (\alpha_{i_1^*j_1^*} - \alpha_{i_2^*j_2^*} ) P [h(\mathbb{D}_{H,S_{j_2^*}})]\right|  \\ 
&  + \left|(1-\alpha_{i_1^*j_1}^*) P [h(\mathbb{D}_{G_{i_1^*},S_{j_1^*}})]  -
(1-\alpha_{i_1^*j_1^*}) P [h(\mathbb{D}_{G_{i_2^*},S_{j_2^*}})] + (\alpha_{i_2^*j_2^*}-\alpha_{i_1^*j_1^*}) P [h(\mathbb{D}_{G_{i_2^*},S_{j_2^*}})]  \right| \bigg\} \\
& \leq 2 \sup_{h\in\mathcal{H}} \bigg\{ \left| 
\alpha_{i_1^*j_1^*} P [h(\mathbb{D}_{H,S_{j_1^*}})]  -
\alpha_{i_1^*j_1^*} P [h(\mathbb{D}_{H,S_{j_2^*}})] \right| + |\alpha_{i_1^*j_1^*} - \alpha_{i_2^*j_2^*}| \\ 
& ~~~~~~~~~~~~~~~ + \left|(1-\alpha_{i_1^*j_1^*}) P [h(\mathbb{D}_{G_{i_1^*},S_{j_1^*}})]  -
(1-\alpha_{i_1^*j_1^*}) P [h(\mathbb{D}_{G_{i_2^*},S_{j_2^*}})] \right| + |\alpha_{i_1^*j_1^*} - \alpha_{i_2^*j_2^*}|\bigg\} \\
& \leq 2 \sup_{h\in\mathcal{H}} \bigg\{ \left| 
\alpha_{i_1^*j_1^*} P [h(\mathbb{D}_{H,S_{j_1^*}})]  -
\alpha_{i_1^*j_1^*} P [h(\mathbb{D}_{H,S_{j_2^*}})] \right| \bigg\} \\ 
& ~~~~ + 2\sup_{h\in\mathcal{H}} \bigg\{ \left|(1-\alpha_{i_1^*j_1^*}) P [h(\mathbb{D}_{G_{i_1^*},S_{j_1^*}})]  -
(1-\alpha_{i_1^*j_1^*}) P [h(\mathbb{D}_{G_{i_2^*},S_{j_2^*}})] \right| \bigg\} + 4|\alpha_{i_1^*j_1^*} - \alpha_{i_2^*j_2^*}| \\
&= \alpha_{i_1^*j_1^*} d_\mathcal{H} (\mathbb{D}_{H,S_{j_1^*}},\mathbb{D}_{H,S_{j_2^*}} ) + 
(1-\alpha_{i_1^*j_1^*}) d_\mathcal{H} (\mathbb{D}_{G_{i_1^*},S_{j_1^*}},\mathbb{D}_{G_{i_2^*},S_{j_2^*}}) + 4|\alpha_{i_1^*j_1^*} - \alpha_{i_2^*j_2^*}| \\ 
&\leq d_\mathcal{H} (\mathbb{D}_{H,S_{j_1^*}},\mathbb{D}_{H,S_{j_2^*}} ) + 
d_\mathcal{H} (\mathbb{D}_{G_{i_1^*},S_{j_1^*}},\mathbb{D}_{G_{i_2^*},S_{j_2^*}}) + 4|\alpha_{i_1^*j_1^*} - \alpha_{i_2^*j_2^*}| \\ 
&\leq \max_{a,b} {d}_\mathcal{H} (\mathbb{D}_{H,S_a},\mathbb{D}_{H,S_b} ) + \max_{a,b,c,d} {d}_\mathcal{H} (\mathbb{D}_{G_c,S_a},\mathbb{D}_{G_d,S_b} )  + 4 \max_{\alpha_{i_1j_1},\alpha_{i_2j_2}} \left| \alpha_{i_1j_1} - \alpha_{i_2j_2} \right |. \\ 
\end{aligned}
\end{equation}
\textit{The second step follows from the triangle inequality. 
The fourth step is by definition.} 

Built on Lemmas ~\ref{lemma:1} and ~\ref{lemma:2}, we have the following lemma.

\vspace{1em}
\noindent\textbf{Lemma 3.} \otherlabel{theorem:0}{3}
\textit{The following bound for $\epsilon_u(h)$ holds: }
\begin{equation}
\label{eq:app_theorem}
\begin{aligned}
\epsilon_u(h) \leq &\sum_{ij} \pi^*_{ij} \epsilon_{ij}(h)+ 
\frac{1}{2} \max_{a,b} {d}_\mathcal{H} (\mathbb{D}_{H,S_a},\mathbb{D}_{H,S_b} ) + \frac{1}{2} \max_{a,b,c,d} {d}_\mathcal{H} (\mathbb{D}_{G_c,S_a},\mathbb{D}_{G_d,S_b} ) + C_0, \\
\end{aligned}
\end{equation}
\begin{equation}
\label{eq:app_constant}
\begin{aligned}
C_0 = \frac{\gamma}{2} + 
\epsilon^* + 2 \max_{\alpha_{i_1j_1},\alpha_{i_2j_2}} \left| \alpha_{i_1j_1} - \alpha_{i_2j_2} \right |. 
\end{aligned}
\end{equation}

\textit{\textbf{Proof.} By Lemma~\ref{lemma:1}, we have
}

\begin{equation}
\label{eq:proof_theorem1}
\begin{aligned}
\epsilon_u(h) & \leq \sum_{ij} \pi^*_{ij} \epsilon_{ij}(h)+ 
\frac{\gamma + \rho}{2}  + \epsilon^* \\
& = \sum_{ij} \pi^*_{ij} \epsilon_{ij}(h)+ 
\frac{\rho}{2}  + \frac{\gamma}{2} + \epsilon^* \\
& \leq \sum_{ij} \pi^*_{ij} \epsilon_{ij}(h)+ 
\frac{1}{2} \max_{a,b} {d}_\mathcal{H} (\mathbb{D}_{H,S_a},\mathbb{D}_{H,S_b} ) + 
\frac{1}{2} \max_{a,b,c,d} {d}_\mathcal{H} (\mathbb{D}_{G_c,S_a},\mathbb{D}_{G_d,S_b})  + C_0. \\ 
\end{aligned}
\end{equation}
\textit{The last step directly follows from Lemma~\ref{lemma:2}.}






The following lemma connects $\mathcal{H}$-divergence and its empirical estimation in the context of our study.

\vspace{1em}
\noindent\textbf{Lemma 4.} \otherlabel{lemma:4}{4} 
\textit{Let $d_{VC}$ be the VC dimension of a hypothesis space $\mathcal{H}$. We denote $N_1$ as the minimum number of human-written or LLM-generated training documents in a domain. 
The following inequality holds with probability at least $(1-\eta)^2$ for any $\eta \in (0,1)$.}
\begin{equation}
\begin{aligned}
   \frac{1}{2} {d}_\mathcal{H} (\mathbb{D}_{H,S_e},\mathbb{D}_{H,S_f}) + 
    \frac{1}{2} {d}_\mathcal{H} (\mathbb{D}_{G_c,S_a},\mathbb{D}_{G_d,S_b})
    & \leq 
    \frac{1}{2} \hat{d}_\mathcal{H} (\mathcal{D}_{H,S_e},\mathcal{D}_{H,S_f}) + \frac{1}{2}  \hat{d}_\mathcal{H} (\mathcal{D}_{G_c,S_a},\mathcal{D}_{G_d,S_b}) \\
    &+ 4 \sqrt{\frac{ d_{VC} \log (2N_1)+\log(\frac{2}{\eta})}{N_1}},
\end{aligned}
\end{equation}
\textit{where domain IDs $a, b ,e ,f \in \{1,2,\dots,n\}$ and LLM IDs $c, d\in \{1,2,\dots,m\}$.}


\textit{\textbf{Proof.}} \textit{We begin the proof by introducing Theorem 3.4 in \citetec{kifer2004detecting2} with a slight adaptation to the context of our study. Let $\mathcal{D}'$ and $\mathcal{D}''$ be two datasets drawn from distributions $\mathbb{D}'$ and $\mathbb{D}''$,  respectively. We denote $N'$ and $N''$ as the number of instances in the datasets $\mathcal{D}'$ and $\mathcal{D}''$, respectively. The following inequality holds:}
\begin{equation}
\begin{aligned}
\label{eq:kifer}
    P \left[ \left| d_\mathcal{H} (\mathbb{D}', \mathbb{D}'') - \hat{d}_\mathcal{H} (\mathcal{D}', \mathcal{D}'') \right|
    > \epsilon \right ] \leq (2N')^{d_{VC}}e^{-N'\epsilon^2/16} + (2N'')^{d_{VC}}e^{-N''\epsilon^2/16}.
\end{aligned}
\end{equation}

\textit{Let $f(N)=(2N)^{d_{VC}}e^{-N\epsilon^2/16}$. Then $f'(N)=f(N)(\frac{d_{VC}}{N}-\frac{\epsilon^2}{16})$. Asymptotically, if $N$ is large enough when $N>\frac{16d_{VC}}{\epsilon^2}$, $f(N)$ is monotonically decreasing. Without loss of generality, let $N'\leq N''$ and the Inequality~\ref{eq:kifer} becomes }
\begin{equation}
    P \left[ \left|d_\mathcal{H} (\mathbb{D}', \mathbb{D}'') - \hat{d}_\mathcal{H} (\mathcal{D}', \mathcal{D}'') \right|
    > \epsilon \right ] \leq 2 (2N')^{d_{VC}}e^{-N'\epsilon^2/16}.
\end{equation}

\textit{Let $\eta = 2(2N')^{d_{VC}}e^{-N'\epsilon^2/16}$. For any $\eta\in(0,1)$, the following inequality holds with probability at least $(1-\eta)$}
\begin{equation}
\begin{aligned}
\label{eq:a13}
    d_\mathcal{H} (\mathbb{D}', \mathbb{D}'') \leq  \hat{d}_\mathcal{H} (\mathcal{D}', \mathcal{D}'') 
     + 4  \sqrt{\frac{ d_{VC} \log (2N')+\log(\frac{2}{\eta})}{N'}} .
\end{aligned}
\end{equation}

\textit{Applying the relationship between $\mathcal{H}$-divergence and its empirical estimation established in Inequality \ref{eq:a13} to our problem context, we have the following inequalities that hold with probability at least $(1-\eta)$ for any $\eta\in(0,1)$:}
\begin{equation}
\begin{aligned}
\label{eq:lemma4-1}
    \frac{1}{2} d_\mathcal{H} (\mathbb{D}_{H,S_e},\mathbb{D}_{H,S_f}) 
    & \leq 
    \frac{1}{2} \hat{d}_\mathcal{H} (\mathcal{D}_{H,S_e},\mathcal{D}_{H,S_f}) + 2 \sqrt{\frac{ d_{VC} \log (2 N_1)+\log(\frac{2}{\eta})}{ N_1}}, \\
\end{aligned}
\end{equation}
\begin{equation}
\begin{aligned}
\label{eq:lemma4-2}
    \frac{1}{2}  {d}_\mathcal{H} (\mathbb{D}_{G_c,S_a},\mathbb{D}_{G_d,S_b}) 
    \leq \frac{1}{2}  {d}_\mathcal{H} (\mathcal{D}_{G_c,S_a},\mathcal{D}_{G_d,S_b}) + 2 \sqrt{ \frac{d_{VC} \log (2 N_1)+\log(\frac{2}{\eta})}{ N_1}}, 
\end{aligned}
\end{equation}
\textit{where $N_1$ is the minimum number of human-written or LLM-generated training documents in a domain.} 
\textit{Given Inequalities~\ref{eq:lemma4-1} and ~\ref{eq:lemma4-2}, their summation is also bounded. Therefore, for any $\eta\in (0,1)$, with probability at least $(1-\eta)^2$, }
\begin{equation}
\begin{aligned}
   \frac{1}{2} {d}_\mathcal{H} (\mathbb{D}_{H,S_e},\mathbb{D}_{H,S_f}) + 
    \frac{1}{2} {d}_\mathcal{H} (\mathbb{D}_{G_c,S_a},\mathbb{D}_{G_d,S_b})
    & \leq 
    \frac{1}{2} \hat{d}_\mathcal{H} (\mathcal{D}_{H,S_e},\mathcal{D}_{H,S_f}) + \frac{1}{2}  \hat{d}_\mathcal{H} (\mathcal{D}_{G_c,S_a},\mathcal{D}_{G_d,S_b}) \\
    &+ 4 \sqrt{\frac{ d_{VC} \log (2N_1)+\log(\frac{2}{\eta})}{N_1}}. 
\end{aligned}
\end{equation}


\vspace{1em}
\noindent\textbf{Theorem 1.}  
\textit{For any $\eta\in(0,1)$, with probability at least $(1-\eta)^2$, the following bound for $\epsilon_u(h)$ holds: }
\begin{equation}
\label{eq:app_prop1}
\begin{aligned}
\epsilon_u(h) \leq &\sum_{ij} \pi^*_{ij} \epsilon_{ij}(h)+ 
\frac{1}{2} \max_{a,b} \hat{d}_\mathcal{H} (\mathcal{D}_{H,S_a},\mathcal{D}_{H,S_b} ) + \frac{1}{2} \max_{a,b,c,d} \hat{d}_\mathcal{H} (\mathcal{D}_{G_c,S_a},\mathcal{D}_{G_d,S_b} ) + C, \\
\end{aligned}
\end{equation}
\begin{equation}
\label{eq:app_propconstant}
\begin{aligned}
C = \frac{\gamma}{2} + \epsilon^* + 2 \max_{\alpha_{i_1j_1},\alpha_{i_2j_2}} \left| \alpha_{i_1j_1} - \alpha_{i_2j_2} \right | + 4 \sqrt{\frac{ d_{VC} \log (2N_1)+\log(\frac{2}{\eta})}{N_1}}. 
\end{aligned}
\end{equation}

\textit{\textbf{Proof.} Theorem~\ref{proposition:1} follows by applying the relationship between $\mathcal{H}$-divergence and its empirical estimation established in Lemma~\ref{lemma:4} and substituting the two $\mathcal{H}$-divergences in Lemma~\ref{theorem:0} with their empirical estimations. The terms in $C$ either attain their respective minimum values (i.e., $\gamma$ and $\epsilon^{*}$) or are constant (i.e., the last two terms in $C$). 
}



\newpage

\section{IMPLEMENTATION DETAILS OF DETECTION METHODS}
\label{appendix:implementation-details}
Our proposed method was trained using the Adam optimizer~\citepec{adam2} with a learning rate of $5 \times 10^{-5}$ for four epochs. The number of memory units $Q$ in a memory bank was set to $10$, and each memory unit was implemented as a $768$-dimensional vector. The hyperparameters $r_1$ and $r_2$ in Equation \ref{eq:kernal} were set to $-3$ and $1$, respectively. The weights $\lambda_y$, $\lambda_h$, and $\lambda_g$ in Equation \ref{eq:loss_all} were set to  $0.1$, $0.2$, and $0.2$, respectively. 
Zero-shot methods, DetectGPT and Fast-DetectGPT, employed the T5-3B model~\citepec{raffel2020exploring2} to perturb texts and utilized the GPT-NeoX model as the scoring LLM. 
Another zero-shot method, DetectLLM, also used the GPT-NeoX model as the scoring model. 
Feature-based methods, GLTR and the multi-feature method, trained a logistic regression model using probability-based or linguistic features extracted from the training dataset.
We used the off-the-shelf OpenAI detector publicly available on Hugging Face.\SingleSpacedXII\footnotemark\footnotetext{See \url{https://huggingface.co/openai-community/roberta-large-openai-detector} (last accessed on June 15, 2025)}\DoubleSpacedXII~ Other fine-tuning benchmarks, EAGLE and DaTeD, employed the same DistilRoBERTa backbone \citepec{sanh2019distilbert2} as our method. 

\bibliographystyleec{apalike} 
\bibliographyec{reference_ec} 

\end{appendices}

\end{document}